\documentclass[letterpaper]{article}
\usepackage[preprint]{aaai2027}
\usepackage[hyphens]{url}
\usepackage{graphicx}
\usepackage{natbib}
\usepackage{caption}
\usepackage{algorithm}
\usepackage{algorithmic}
\usepackage{amssymb}
\usepackage{amsmath}

\usepackage{newfloat}
\usepackage{listings}
\DeclareCaptionStyle{ruled}{labelfont=normalfont,labelsep=colon,strut=off}
\floatstyle{ruled}
\newfloat{listing}{tb}{lst}{}
\floatname{listing}{Listing}

\usepackage{booktabs}
\usepackage{array}

\title{V-Mem: Modality-Routed Retrieval for Long-Term Multimodal Agentic Memory}

\author{
    Dingyi Kang\textsuperscript{\rm 1},
    Dongming Jiang\textsuperscript{\rm 1},
    Yi Li\textsuperscript{\rm 1},
    Guanpeng Li\textsuperscript{\rm 2},
    Bingzhe Li\textsuperscript{\rm 1}
}
\affiliations{
    \textsuperscript{\rm 1}Department of Computer Science, The University of Texas at Dallas\\
    \textsuperscript{\rm 2}Department of Electrical and Computer Engineering, University of Florida\\
    \{dingyi.kang, dongming.jiang, yi.li3, bingzhe.li\}@utdallas.edu, liguanpeng@ufl.edu
}

\begin{document}
\maketitle

\begin{abstract}
Interaction between users and LLM agents is increasingly multimodal: conversations interleave text with images, and a later question may target either. Yet most agent memories are designed around text, and even the few that support multimodal conversations still fail on vision-related questions. We trace this failure to an assumption behind the \emph{similarity search} they rely on: in the index space, a query lies close to the relevant evidence that answers it. In multimodal settings, two gaps break it. By the \emph{modality gap}, a query lies closer to memory content of its own modality than to evidence in another, even in a trained joint embedding space. By the \emph{similarity-relevance gap}, the content most similar to a query is often not the evidence that answers it, most acutely when a query carries both text and image and its evidence resembles neither part alone. We present V-Mem, a multimodal agentic memory system that routes retrieval by the modality of the query and that of the target evidence, both recognized from the query alone. To cross the modality gap, V-Mem organizes the conversation into rounds and returns the target-modality content from the same round as the match, without comparing across modalities. To close the similarity-relevance gap, it searches with an LLM-generated anchor that sits closer to the relevant evidence than the query does: a hypothetical caption for a text-only query seeking an image, and an enriched search anchor, the query text plus relevant keywords extracted from the query image, when the evidence is reachable only by combining the two. On Mem-Gallery, V-Mem reaches an LLM-judge score of 0.82 versus 0.56 for the second best, with the largest margin on questions carrying an image (0.87, no baseline above 0.47); on LoCoMo it scores 0.69 versus 0.58.
\end{abstract}

\begin{links}
    \link{Code}{https://github.com/Dingyi-Kang/V-Mem}
\end{links}

\section{Introduction}
\label{sec:intro}

\begin{figure}[t!]
\centering
\includegraphics[width=0.97\linewidth]{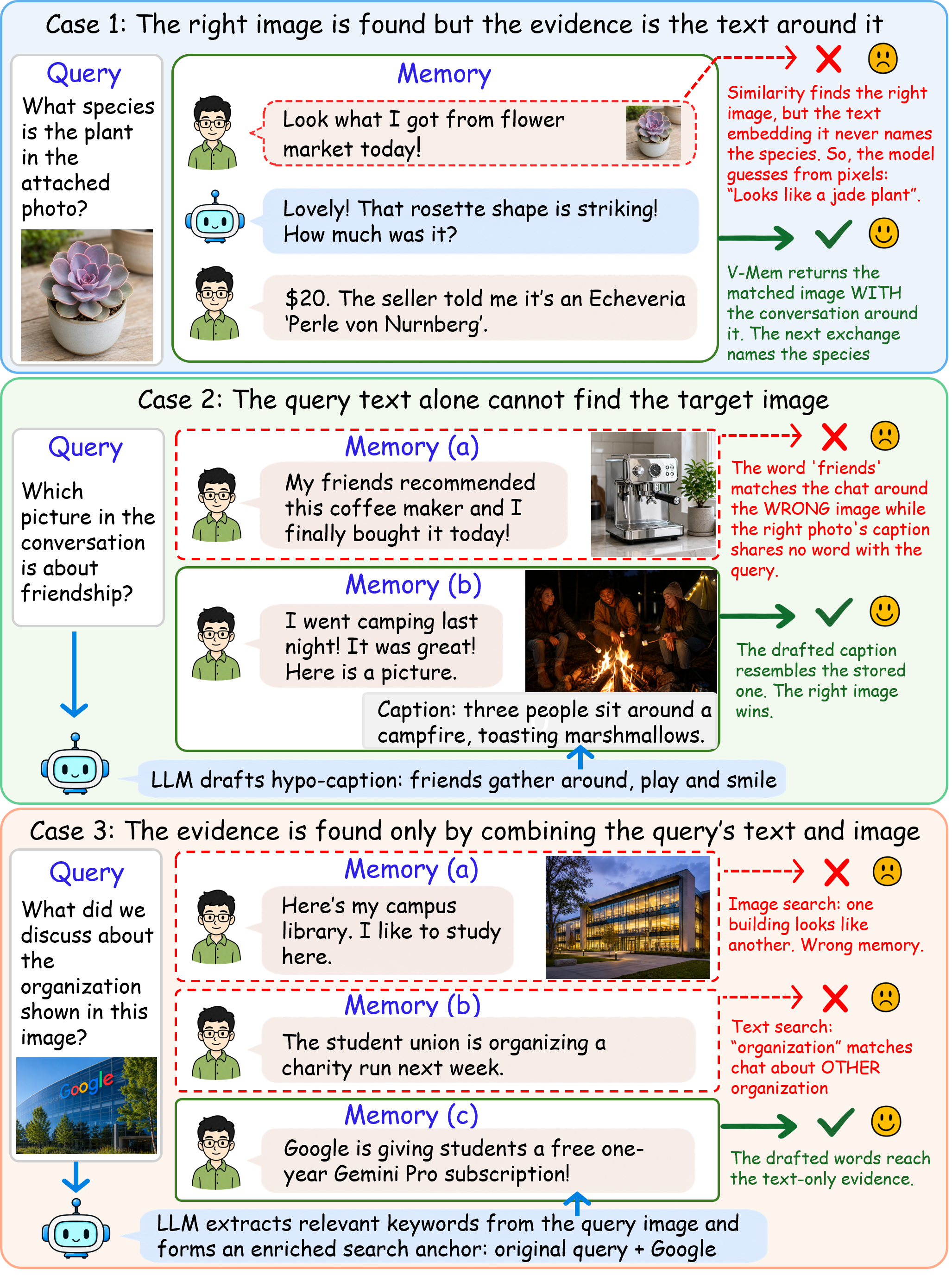}
\caption{Three cases where similarity search fails (red) and V-Mem succeeds (green): (1) the right image is found but the evidence is the text around it; (2) the query text alone cannot find the target image; (3) the evidence is found only by combining the query's text and image. Examples are illustrative.}
\label{fig:motivation}
\end{figure}

Large language models (LLMs) now underpin AI agents that plan, call tools, and interpret both text and images \citep{yao2023react,achiam2023gpt4}. Deployed as long-running assistants, these agents accumulate weeks or months of conversation in which a user's pictures, voice messages, and text messages are tightly interleaved. Retaining this history, however, remains difficult. An LLM operates only over the contents of its current context window, and its internal representations do not persist once earlier content is displaced; information introduced early in an interaction is therefore unavailable when it is later required. Simply enlarging the window does not resolve this: a model attends to long inputs unevenly, and evidence sitting far from the query is weighed only weakly, effects documented as \emph{lost-in-the-middle} and context decay \citep{liu2024lostmiddle,levy2024moretokens}.

Memory-Augmented Generation (MAG) confronts this by placing storage outside the model \citep{packer2023memgpt,memos,amem,magma}. Rather than relying on the context window to carry everything forward, a MAG system writes the interaction history to an external store and, at each new turn, retrieves the fragments that bear on the current request and folds them back into generation. Most current MAG systems focus on text alone. They distill a conversation down to text, whether summaries, graph nodes, or sentence embeddings, and find evidence by comparing text against text \citep{amem,simplemem,magma}. Real interactions are rarely so tidy: users photograph a product, screenshot an error, share a diagram, or send a voice message, and later expect the agent to recall it together with whatever was said about it \citep{memgallery}. As multimodal agents become commonplace, a memory that can index only words leaves much of the exchange unrecorded, pushing the question of how to organize and retrieve over conversations that interleave text, images, and audio to the foreground.

A few systems do target multimodal memory \citep{memverse,m2a,omni}, yet they still fail on vision-related questions. The root cause is an assumption behind the similarity search they rely on: in the index space, a query lies close to the relevant evidence that answers it. However, in multimodal settings this assumption is broken by two gaps that push a query and its relevant evidence apart. First, the \emph{modality gap} \citep{liang2022modalitygap}: when a query and its evidence are in different modalities, the query lies closer to memory contents of its own modality than to the evidence, even in a trained joint embedding space, which causes similarity search to rank same-modality content above the evidence. Second is the \emph{similarity-relevance gap} \citep{furnas1987vocabulary,gao2023hyde}: the most similar content in the memory may not be the most relevant evidence for a query, especially in multimodal settings. When a query carries both text and image and its relevant evidence resembles neither individually, finding it requires combining both. Therefore, searching each modality alone retrieves only its most similar content, and causes similarity search to fail to retrieve it.

Consider a user who attaches a photo of the Google headquarters and asks, \emph{``What did we discuss about the organization shown in this image?''} (Case~3 in Figure~\ref{fig:motivation}). The relevant evidence is from an earlier session, \emph{``Google is giving students a free one-year Gemini Pro subscription,''} which resembles neither the query's photo nor its text and thus causes similarity search to fail to retrieve it. Against the photo, the evidence is in a different modality, and the embedding of a building lies nowhere near that of a discussion about a subscription, so searching by the query image retrieves other building photos, but never the text evidence. Against the text, the evidence shares no words with the query, which names neither ``Google'' nor ``subscription'' but only ``organization,'' so a lexical or embedding match on the query text retrieves other memories that mention organizations, not this evidence about the Google subscription.

To close both gaps, we propose V-Mem, a multimodal agentic memory system that routes retrieval by the modality of the query and that of the target evidence, both recognized from the query alone via a keyword rule. Because not every query needs to search over all stored text and images, and uniform search would inject distracting signals that out-rank the relevant evidence, V-Mem keeps a separate retrieval lane per memory modality, matching text against text and images against images, and activates only the lanes a case needs. When the query and its target evidence share a modality, only that lane is activated. When they are in different modalities, both lanes are activated, and V-Mem narrows each gap in turn. To cross the modality gap, V-Mem organizes the conversation into rounds, each a user-assistant exchange; since the content shared within a round is likely relevant to one another, it searches within the lane of the query's modality, finds the top match, and returns the associated target-modality content from the same round as the match, so it is able to retrieve relevant evidence without comparing across modalities. To close the similarity-relevance gap, V-Mem prompts an LLM to generate a search anchor closer to the relevant evidence than the query itself: a \emph{hypothetical caption} for a text-only query seeking an image, and an \emph{enriched search anchor}, the query text plus relevant keywords extracted from the query image, when the evidence is reachable only by combining the two. The experimental results show that V-Mem outperforms multimodal agentic memory baselines on the Mem-Gallery and LoCoMo benchmarks. On Mem-Gallery \citep{memgallery}, it reaches an LLM-judge score of 0.82 versus 0.56 for the second best, with the largest margin on questions that carry an image (0.87, no baseline above 0.47); on LoCoMo \citep{locomo} it scores 0.69 versus 0.58 for the second best. Beyond accuracy, V-Mem builds in seconds with zero LLM tokens.

\section{Background and Problem Formulation}
\label{sec:background}

Memory is a core capability of LLM agents \citep{zhang2025d}: as an interaction outgrows the context window, the agent offloads it to an external store and retrieves from it when answering later queries, rather than encoding everything in model parameters. Prior work terms this Memory-Augmented Generation (MAG) \citep{memos,jiang2026anatomy}.

Formally, a MAG system maintains a memory $\mathcal{M}_t$: at turn $t$ it answers a query $q_t$ from retrieved memory and ingests new content $x_t$,
\begin{equation}
a_t = \mathrm{LLM}\big(q_t,\ \mathcal{R}(q_t, \mathcal{M}_t)\big),
\qquad
\mathcal{M}_{t+1} = \mathcal{U}\big(\mathcal{M}_t,\ x_t\big),
\label{eq:mag}
\end{equation}
where $\mathcal{R}$ retrieves the evidence relevant to $q_t$ from $\mathcal{M}_t$, and $\mathcal{U}$ writes $x_t$ into memory. Almost all systems implement $\mathcal{R}$ as \emph{similarity search}, whether by dense-embedding cosine or sparse lexical matching (BM25), ranking stored memory content by its similarity to the query,
\begin{equation}
\mathcal{R}(q,\mathcal{M}) = \mathop{\mathrm{top}\mbox{-}k}_{u \,\in\, \mathcal{M}}\ s\big(e(q),\, e(u)\big),
\label{eq:simsearch}
\end{equation}
where $e(\cdot)$ maps a query or memory item to a representation, $s(\cdot,\cdot)$ scores their similarity, and the query $q$ serves as the search \emph{anchor}. Either kind of index succeeds only if the \emph{relevant evidence} $u^{\star}$, the content in memory that answers $q$, lies close to the anchor in the index space and so ranks among the top $k$.

\begin{figure*}[t]
\centering
\includegraphics[width=\textwidth]{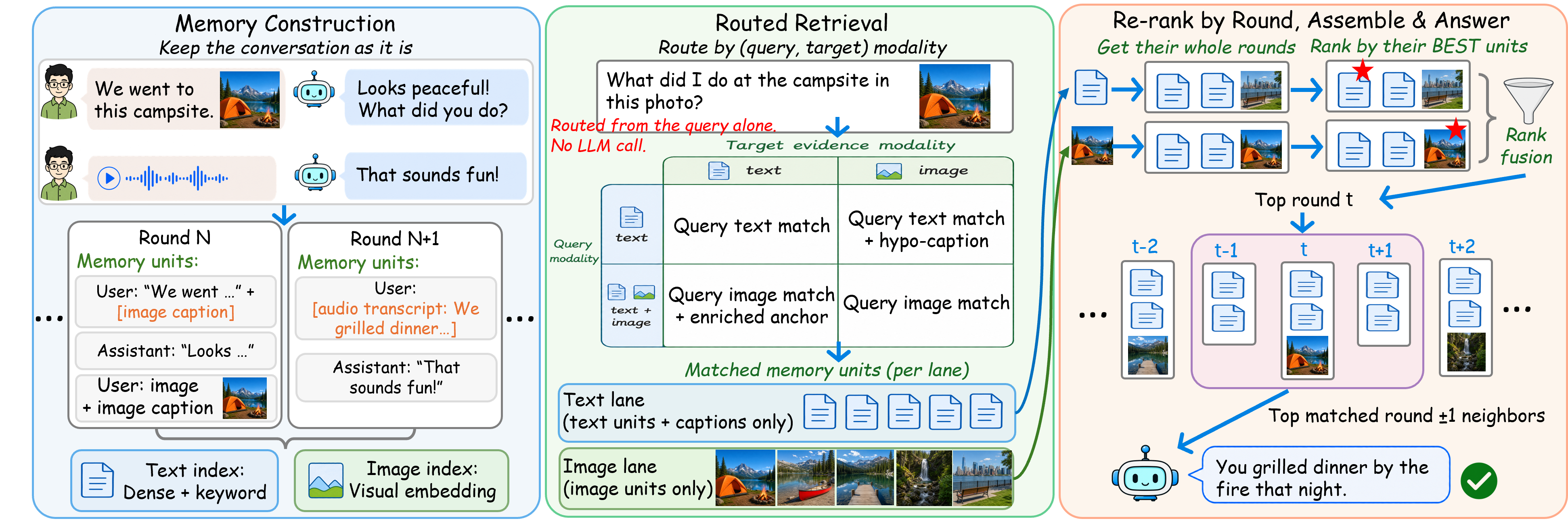}
\caption{The V-Mem pipeline. \emph{Memory construction} keeps the conversation as it is, organizing it into rounds; each round, one user-assistant exchange, holds memory units, each a text span or image from user or assistant. All text, including image captions, is indexed by a dense text embedding and a sparse BM25 index, and the raw image by a visual embedding. \emph{Routed retrieval} recognizes the (query, target) modality pair from the query alone and fires only the lanes that case needs, searching with the appropriate anchor: the raw query, a hypo-caption, or an enriched anchor. \emph{Re-rank by round} scores each round by its best-matching unit, fuses the lanes by rank, and returns each top round with its $\pm1$ neighbors.}
\label{fig:pipeline}
\end{figure*}

As LLMs increasingly handle images and audio alongside text, interactions become multimodal. However, most MAG systems are still built around text \citep{packer2023memgpt,amem,simplemem,magma}; MemVerse \citep{memverse} and Omni-SimpleMem \citep{omni} index images only through captions, so visual content never enters the search. Some systems instead retrieve across modalities directly, matching a text query against stored images or an image query against stored text \citep{m2a,chen2022murag}, but two gaps push the query and its relevant evidence $u^{\star}$ apart and make similarity search fail. First, the \emph{modality gap}: when the query and $u^{\star}$ are in different modalities, the query lies closer to memory content of its own modality than to $u^{\star}$, so similarity search ranks same-modality content above $u^{\star}$ \citep{liang2022modalitygap}; contrastive encoders such as CLIP and SigLIP \citep{clip,siglip} learn a shared space to narrow this, and memory systems such as M2A \citep{m2a} adopt it, but the gap persists even in this trained joint embedding space. Second, the \emph{similarity-relevance gap}: even within one modality, the content in memory most similar to the query is often not $u^{\star}$, because $u^{\star}$ often does not resemble the query it answers \citep{furnas1987vocabulary,gao2023hyde}, especially in the multimodal setting. To narrow this gap, MuRAG \citep{chen2022murag} trains its encoder and retriever together to fetch the most relevant content, but this needs task-specific supervision and does not generalize to general-purpose conversational memory. What is worse, all of these approaches share the same blind spot: a query that carries both text and image, and whose relevant evidence resembles neither part alone. Finding such evidence requires combining the two, yet searching each modality alone retrieves only its most similar content and fails to retrieve $u^{\star}$.

\section{V-Mem Design}
\label{sec:method}

\subsection{Overview}
\label{sec:overview}

V-Mem organizes the interaction history into \emph{rounds}, each one user-assistant exchange, and keeps each text span and image in a round as a separate \emph{unit}. Each unit is indexed and stored in the lane of its modality: text units in a text lane, image units in an image lane (\S\ref{sec:memory}). At query time, rather than apply one uniform search to every query over all memory, V-Mem routes: by a keyword rule it infers the (query, target) modality pair from the query alone, then, based on that pair, decides what to search \emph{over} (which lanes), what to search \emph{with} (the raw query, or an anchor the LLM generates), and what to \emph{return} (which rounds, and whether to add their neighbors). To close the gap between the query and its \emph{target evidence}, the relevant evidence that answers it, V-Mem uses two mechanisms: matching through a shared round to cross the modality gap, and searching with an LLM-generated anchor closer to the target evidence, closing the similarity-relevance gap (\S\ref{sec:retrieval}, Figure~\ref{fig:pipeline}). Because nothing is summarized or rewritten, the memory builds in seconds with no LLM calls (\S\ref{sec:construction}).

\subsection{Rounds, Units, and Lanes}
\label{sec:memory}

V-Mem represents a conversation history as a sequence of rounds $\mathcal{C}=(r_1,\dots,r_T)$, where each round $r_t$ is one user-assistant exchange and holds all the \emph{memory units} (\emph{units} for short) in that exchange: a unit $u$ is a single text span or a single image, with modality $m(u)\in\{\mathsf{txt},\mathsf{img}\}$. The units of a round co-occur in one exchange, so they are likely relevant to one another, which is the premise of matching through shared rounds below. What is more, each image is kept as its own unit, so a query image can be matched against stored images directly.

V-Mem further groups all the units by modality into \emph{lanes}: the text lane holds the text units, and the image lane holds the image units. Two indexes cover these units. The text index carries two complementary signals, a dense embedding from an encoder $\phi$ (semantic) and a sparse BM25 index (lexical), over both the content of text units and the caption text of image units; the image index carries the visual embedding $\psi$ of each raw image. By keeping a separate lane per modality, V-Mem can search only the lanes a query calls for and leave the irrelevant memory untouched.

\subsection{Routed Retrieval}
\label{sec:retrieval}

Routed retrieval generalizes the similarity search of Eq.~(\ref{eq:simsearch}) along three axes: what it searches \emph{over}, what it searches \emph{with}, and what it \emph{returns}. Writing $c=\mathrm{Route}(q)$ for the query's case, retrieval becomes
\begin{equation}
\mathcal{R}_{\text{V-Mem}}(q,\mathcal{M})
= \Pi_c\!\Big(\mathop{\mathrm{top}\mbox{-}k}_{r}\ \mathrm{Fuse}_{\ell\in L_c}\, S_\ell\big(a_{c,\ell}(q),\, \mathcal{U}_{c,\ell}\big)\Big).
\label{eq:routed}
\end{equation}
The case $c$ decides which lanes to activate ($L_c$), and the case together with the lane decides which units within that lane to search over ($\mathcal{U}_{c,\ell}\subseteq\mathcal{M}$) and what to search with ($a_{c,\ell}(q)$, the raw query or an anchor generated by an LLM). Each active lane scores every round independently ($S_\ell$), $\mathrm{Fuse}$ merges the per-lane rankings into one, and $\Pi_c$ assembles the top $k$ rounds into the content returned to the answer LLM. Eq.~(\ref{eq:simsearch}) is the special case of a single lane whose candidate set is all of memory, whose anchor is the raw query, and whose return map is the identity. The rest of this subsection specifies how V-Mem recognizes the case and what each case fixes (full procedure in Algorithm~\ref{alg:vmem}, Appendix~\ref{sec:appendix_impl}).

\paragraph{Routing.}
V-Mem routes each query by its (query, target) modality pair, activating only the lanes and strategy that case needs. It recognizes both modalities from the query alone, with no LLM call: the query modality is decided by whether an image is attached in the query, and the target modality is inferred by a keyword rule that classifies the query as asking for an image or for a fact (full rule in Appendix~\ref{sec:appendix_impl}). Given that a query always carries the question text itself, the modality of the query is either pure text or text with an image, never image alone. In contrast, its target evidence is either text or image. This gives four cases (Figure~\ref{fig:pipeline}): two same-modality retrievals (text$\to$text and image+text$\to$image) and two cross-modality retrievals (text$\to$image and image+text$\to$text). Routing this way keeps a lane from adding distractors to a case it doesn't serve: activating the image lane would distract a text$\to$text retrieval, and vice versa. Keeping separate lanes and routing by modalities enable V-Mem to apply the strategy each case calls for.

Among the four cases, the two same-modality retrievals are direct: the query and its target evidence are in the same modality, so only the lane of the shared modality is activated. The two cross-modality retrievals, text$\to$image and image+text$\to$text, however, face both the modality gap and the similarity-relevance gap at once. To address them, V-Mem activates both lanes and combines two mechanisms: matching through a shared round, and searching with an LLM-generated anchor. We explain each in turn below.

\paragraph{Matching through shared rounds.}
Comparing a query directly against evidence in another modality is unreliable, and thus a way to reach that modality without cross-modal comparison is needed. V-Mem crosses the modality gap through rounds: it matches the query within the lane of its own modality, finds the top match, and returns the target-modality units from the same round as the match. Units in the same round co-occur and tend to be related, so the returned units are likely relevant to the query. For example, in Case~1 of Figure~\ref{fig:motivation}, the matched plant photo cannot name its species, but the exchange beside it does. V-Mem applies this matching in both cross-modality cases:
\begin{itemize}
\item \emph{Text\,$\to$\,image.} The query is text, so V-Mem matches in the text lane, restricted to the text of image-bearing rounds: searching all text would let rounds without images out-rank those that hold the target image, and a round without an image has nothing to return. It then returns the images in the top rounds.
\item \emph{Image+text\,$\to$\,text.} The query carries an image, so V-Mem matches in the image lane, comparing the query image with stored images by similarity between their visual embeddings. It then returns the text of the top rounds.
\end{itemize}
Every match stays within one modality, so V-Mem needs no learned cross-modal alignment; the round, not a shared embedding space, links the two modalities.

\paragraph{Searching with an LLM-generated anchor.}
In multimodal settings, the raw query often does not resemble the target evidence, so matching with it misses that evidence. What is needed is a search anchor that sits closer to the relevant evidence than the query does, and V-Mem obtains one by generating it with an LLM. It takes a different form in each cross-modality case (prompts in Appendix~\ref{sec:appendix_prompts}):
\begin{itemize}
\item \emph{Text\,$\to$\,image.} The query asks to retrieve a stored picture without providing any reference image. V-Mem therefore \emph{replaces} the query with an LLM-generated hypothetical caption (\emph{hypo-caption}), a caption the target image would plausibly have, and matches it against the stored images' captions (Case~2 of Figure~\ref{fig:motivation}). Given that a hypo-caption describes the one image the query is looking for, V-Mem first checks whether the query identifies an image or counts images, based on a keyword rule with no LLM call; a counting query keeps the raw query text, because there is no single image for a caption to describe.
\item \emph{Image+text\,$\to$\,text.} The target evidence often resembles neither the query text nor the attached image alone, so finding it requires both. V-Mem therefore prompts a vision LLM to extract, from the attached image, the keywords the answering turns are likely to contain, and appends them to the query text. The text lane then searches with this \emph{enriched anchor}, which sits closer to the target evidence than the query text or the image alone (Case~3 of Figure~\ref{fig:motivation}).
\end{itemize}

\paragraph{Scoring and fusion.}
Each active lane carries one or more \emph{signals}, the ways its units can be scored. The text lane scores its text with a dense embedding ($\phi$) and a sparse BM25 index. The image lane scores an image either by the visual embedding ($\psi$) of the raw image, when the query carries an image, or by the dense and sparse signals over the image's caption, when it does not. Let $J_c$ collect the signals that case $c$ activates. A signal $j\in J_c$ in lane $\ell$ scores the units of $\mathcal{U}_{c,\ell}$ against that lane's anchor $a_{c,\ell}(q)$ using its own encoder $e_j$: the dense and visual signals take cosine similarity,
\begin{equation}
s_j(u)=\cos\!\big(e_j(a_{c,\ell}),\, e_j(u)\big),
\label{eq:lane}
\end{equation}
and the sparse signal uses BM25. Scores are computed per unit, but retrieval ranks rounds, so under each signal a round takes the highest score of its units. Because the signals sit on different scales (bounded cosine versus unbounded BM25), V-Mem fuses them by \emph{rank} rather than raw score \citep{rrf}:
\begin{equation}
S_j(r)=\max_{u\in r_\ell}\, s_j(u),
\qquad
\mathrm{score}(r)=\sum_{j\in J_c}\frac{1}{\kappa+\mathrm{rank}_j(r)},
\label{eq:fuse}
\end{equation}
where $r_\ell = r \cap \mathcal{U}_{c,\ell}$ collects the units of round $r$ that lane $\ell$ ranks, $S_j(r)$ is the score round $r$ earns under signal $j$, $\mathrm{rank}_j(r)$ is its position when all rounds are sorted by that score, and $\kappa$ is a smoothing constant. Ranking puts the signals on a common footing, so no signal dominates by scale. Fusion returns the top-$k$ rounds under $\mathrm{score}(r)$.

\paragraph{Context assembly.}
What V-Mem returns depends on the query type. For most queries, it wraps each top round with its $\pm1$ adjacent rounds into one time-ordered span: adjacent rounds tend to develop one topic, so a match's neighbors often complete its evidence, from a follow-up detail to the next hop of a reasoning chain. These adjacent rounds serve a similar role to the links among entities in graph-based memories. Aggregation questions call for the opposite: their evidence is scattered across many distant rounds. For example, \emph{``how many different museums have we discussed?''} draws on the whole history. Neighbors add little in this case, and within a limited return budget they would crowd out the many rounds such questions need, so for such queries V-Mem drops the neighbors and returns more rounds instead. V-Mem recognizes these questions from the query's wording by a keyword rule, with no LLM call (full rule in Appendix~\ref{sec:appendix_impl}).

\subsection{Memory Construction}
\label{sec:construction}

V-Mem builds its memory with local encoders alone, keeping the conversation as it is rather than rewriting it. Extraction-based construction spends LLM calls to compress the history into summaries, notes, or graph nodes, and whatever it drops is unavailable at retrieval time; V-Mem keeps every span and image and leaves the selection to query time. For each round, it emits one text unit per span and one image unit per image, then indexes each unit and keeps it in its lane (\S\ref{sec:memory}). Audio is treated as text: a voice message is transcribed and stored as a text unit of its round (as ``\texttt{[Audio: \textless transcript\textgreater]}''), searchable through the text lane with no dedicated audio index.

Figure~\ref{fig:pipeline} (left) traces this on two exchanges. In the first, the user sends a message with a photo of a campsite and the assistant replies: the round holds the user's text span, the assistant's text span, and the photo itself, and the photo's caption joins the text index so the photo can also be reached by its description. In the second, the user sends a voice message, whose transcript becomes a text unit of that round.

Moreover, this memory construction is incremental: the update operator $\mathcal{U}$ of Eq.~(\ref{eq:mag}) appends each new round's units and encodes them locally, leaving the existing memory untouched. Since each round is encoded on its own, the memory can be built as the conversation happens rather than in a batch pass over the history, and it never needs rebuilding as the history grows. Given no summarization or LLM extraction, the memory builds in seconds at zero token cost (\S\ref{sec:efficiency}).

\section{Experiments}
\label{sec:experiments}

\subsection{Setup}
\label{sec:setup}

\paragraph{Benchmarks.}
We evaluate on two long-term conversational-memory benchmarks. Mem-Gallery \citep{memgallery} is multimodal: 20 multi-session dialogues with 1{,}711 questions over nine categories. Three are visual: visual-centric search (VS), visual-centric reasoning (VR), and test-time learning (TTL), in which the query supplies a new image. The other six are text-centric: factual retrieval (FR), temporal reasoning (TR), multi-entity reasoning (MR), knowledge resolution (KR), conflict detection (CD), and answer refusal (AR). LoCoMo \citep{locomo} has 10 conversations with 1{,}986 questions over five categories (single-hop, multi-hop, temporal, open-domain, adversarial); following common practice we evaluate the four non-adversarial ones ($1{,}540$ questions). It gives images only as caption text, which suffices for every image-related question. Mem-Gallery mainly tests retrieval across modalities; LoCoMo tests retrieval within text.

\paragraph{Baselines.}
We compare against three memory systems that differ in how they represent images. Omni \citep{omni} is a graph- and summary-based memory that extracts memory units with an LLM and indexes images only through their captions. M2A \citep{m2a} is an agentic memory that keeps each image as both a caption and an embedding, and retrieves by fusing dense, sparse, and image signals in an iterative LLM-in-the-loop pipeline. A-Mem \citep{amem} is a text-only agentic memory that builds and evolves note abstractions with an LLM per turn.

\begin{table*}[t]
\centering
\caption{LLM-judge scores on Mem-Gallery, by category and overall, under two backbone models.
The judge is \texttt{gpt-4o-mini} throughout, so only the backbone differs between blocks.
Within each block, the best score in each column is bold and the second best underlined.
Lexical metrics are in Appendix~\ref{sec:appendix_lexical}.}
\label{tab:mg_percat}
\small
\begin{tabular}{@{}l rrrrrrrrr r@{}}
\toprule
System & TTL & VR & VS & AR & CD & KR & MR & FR & TR & Overall \\
\midrule
\multicolumn{11}{@{}l}{\textit{\texttt{gpt-4o-mini} backbone}}\\
Omni (SimpleMem) & 0.141 & \underline{0.273} & \underline{0.658} & 0.796 & \underline{0.438} & \underline{0.531} & \underline{0.840} & \underline{0.801} & \underline{0.736} & \underline{0.561} \\
M2A              & \underline{0.472} & 0.106 & 0.016 & 0.736 & 0.383 & 0.210 & 0.447 & 0.203 & 0.276 & 0.314 \\
A-Mem            & 0.163 & 0.261 & 0.319 & \underline{0.802} & 0.364 & 0.488 & 0.820 & 0.758 & 0.301 & 0.460 \\
\textbf{V-Mem (ours)} & \textbf{0.872} & \textbf{0.575} & \textbf{0.833} & \textbf{0.940} & \textbf{0.667} & \textbf{0.698} & \textbf{0.918} & \textbf{0.881} & \textbf{0.785} & \textbf{0.825} \\
\midrule
\multicolumn{11}{@{}l}{\textit{Qwen2.5-VL-32B backbone}}\\
Omni (SimpleMem) & 0.190 & \underline{0.420} & \underline{0.735} & 0.837 & \underline{0.414} & \underline{0.654} & \underline{0.891} & \underline{0.854} & \textbf{0.825} & \underline{0.628} \\
M2A              & \underline{0.540} & 0.118 & 0.033 & 0.867 & 0.278 & 0.080 & 0.296 & 0.123 & 0.012 & 0.287 \\
A-Mem            & 0.273 & 0.382 & 0.449 & \underline{0.875} & 0.358 & 0.549 & 0.847 & 0.774 & 0.337 & 0.535 \\
\textbf{V-Mem (ours)} & \textbf{0.912} & \textbf{0.695} & \textbf{0.869} & \textbf{0.886} & \textbf{0.475} & \textbf{0.710} & \textbf{0.900} & \textbf{0.897} & \underline{0.679} & \textbf{0.829} \\
\bottomrule
\end{tabular}
\end{table*}

\begin{table}[t]
\centering
\caption{LLM-judge scores on LoCoMo, by category and overall. Best per column in bold, second best underlined.}
\label{tab:locomo_percat}
\small
\setlength{\tabcolsep}{4pt}
\begin{tabular}{@{}l rrrr r@{}}
\toprule
System & S-hop & M-hop & Temp. & Open & Overall \\
\midrule
Omni (SimpleMem) & \underline{0.663} & \textbf{0.555} & 0.442 & \underline{0.440} & \underline{0.583} \\
M2A              & 0.491 & 0.363 & 0.274 & \textbf{0.477} & 0.422 \\
A-Mem            & 0.572 & 0.423 & \underline{0.485} & 0.331 & 0.512 \\
\textbf{V-Mem (ours)} & \textbf{0.818} & \underline{0.550} & \textbf{0.574} & 0.375 & \textbf{0.690} \\
\bottomrule
\end{tabular}
\end{table}

\paragraph{Metrics.}
Our primary metric is an LLM-as-judge score (0--1) from \texttt{gpt-4o-mini}, which scores each answer independently against the gold reference, so all systems are directly comparable. Also, we report lexical overlap (e.g., F1, BLEU, and exact match) in Appendix~\ref{sec:appendix_lexical}.

\paragraph{Backbone model.}
We run each baseline using its original implementation and recommended settings. Within each experiment all systems share the same backbone and the same text and image encoders, so differences reflect memory design rather than a stronger backbone or encoder. The main results use \texttt{gpt-4o-mini} and a locally served Qwen2.5-VL-32B backbone; Appendix~\ref{sec:appendix_local7b} adds a smaller Qwen2.5-VL-7B. Every answer is scored by the same \texttt{gpt-4o-mini} judge in all three settings, using Mem-Gallery's official judge prompt and, for LoCoMo, the prompt established by prior work \citep{magma}.

\paragraph{Implementation.}
V-Mem uses \texttt{all-MiniLM-L6-v2} \citep{sbert} for text embeddings, BM25 \citep{bm25} for lexical match, and SigLIP \citep{siglip} for image embeddings, all running locally; the anchors are generated at answer time.

\subsection{Main Results}
\label{sec:main_results}

\paragraph{V-Mem wins the LLM judge on both benchmarks.}
Tables~\ref{tab:mg_percat} and~\ref{tab:locomo_percat} report LLM-judge scores per category and overall. On Mem-Gallery, V-Mem reaches 0.825, against 0.561 for Omni, the strongest baseline, then 0.460 for A-Mem and 0.314 for M2A. On LoCoMo it scores 0.690 against Omni's 0.583. The ordering is unchanged under a local Qwen2.5-VL-32B backbone: $0.829$ for V-Mem, against Omni's $0.628$, A-Mem's $0.535$ and M2A's $0.287$. There the visual margin widens further, to $0.912$ on provided-image questions where no baseline exceeds $0.540$, and Omni leads only on temporal reasoning ($0.825$ versus $0.679$). Under a smaller 7B backbone the ordering does change (Appendix~\ref{sec:appendix_local7b}). Lexical metrics are in Appendix~\ref{sec:appendix_lexical}: V-Mem wins every one on Mem-Gallery, while on LoCoMo they favor Omni's longer, gold-style answers, which lexical overlap rewards regardless of correctness.

\paragraph{The visual categories drive the Mem-Gallery gap.}
The largest margins in Table~\ref{tab:mg_percat} fall exactly where modality-matched evidence matters. On provided-image questions (TTL) V-Mem reaches 0.872 where no baseline exceeds 0.472, and on visual reasoning (VR) 0.575 against at most 0.273. The baselines store images as captions or as coarse cross-modal embeddings, so none can match a query image against the stored images. V-Mem does, and returns the text bound to the match in the same round. The advantage also holds on non-visual categories (CD, KR, MR), so it is not purely a vision effect.

\paragraph{LoCoMo: wins where evidence is retrievable.}
Per category (Table~\ref{tab:locomo_percat}), V-Mem wins single-hop decisively ($0.818$ against $0.663$) and temporal ($0.574$ against $0.485$), and it ties multi-hop ($0.550$ against Omni's $0.555$) without building any graph. Its one shortfall is open-domain, where M2A leads ($0.477$ against $0.375$).

\subsection{Ablations}
\label{sec:ablations}

\begin{table}[t]
\centering
\caption{Incremental ablation of the core design on Mem-Gallery, scored by the LLM judge:
the three visual categories these steps most affect, plus the overall score across all nine.
Best per column in bold.}
\label{tab:abl_incremental}
\small
\setlength{\tabcolsep}{2pt}
\begin{tabular}{@{}l rrr r@{}}
\toprule
Configuration & TTL & VR & VS & Overall \\
\midrule
Vanilla (all lanes, unit-level)         & 0.740 & 0.362 & 0.655 & 0.680 \\
\;\;+ routing                           & 0.708 & 0.356 & 0.647 & 0.689 \\
\;\;+ matching through shared rounds    & \textbf{0.890} & 0.537 & 0.801 & 0.818 \\
\;\;+ searching with a generated anchor & 0.872 & \textbf{0.575} & \textbf{0.833} & \textbf{0.825} \\
\bottomrule
\end{tabular}
\end{table}

\paragraph{Building the design up.}
Table~\ref{tab:abl_incremental} rebuilds V-Mem from a naive multimodal retrieval baseline, one design decision per row, with the return budget held fixed throughout: every row returns the same number of most-relevant items, so each step changes what those items are, never how many. \emph{Vanilla} makes no decisions: every lane the query physically supports fires for every question, one uniform fusion ranks the memory units, and the most relevant units are returned bare. \emph{+ Routing} adds only the $2\times2$ case recognition and lane assignment: same-modality cases keep the other modality's lanes out, cross-modal cases keep both, and every lane still searches with the raw query. \emph{+ Matching through shared rounds} returns the match's whole round rather than bare units. This does more than add surrounding text: a text match can now return the image beside it in the same round, and vice versa. \emph{+ Searching with a generated anchor} lets the lanes search with a hypothetical caption or an enriched search anchor instead of the raw query.

Each step lands where it aims, and each unlocks the next. Routing lifts every text category at once (KR $+0.043$, MR $+0.049$, FR $+0.037$), but leaves the visual categories flat and dips provided-image questions (TTL $0.740 \to 0.708$). The reason for the dip is what routing switches off: Vanilla had fired every lane the query supported, including auxiliary ones that routing drops, and those had been reaching some of the answering text by chance. 37 TTL questions lose it against 20 that gain. As for matching through shared rounds, it supplies that reach by design and is by far the largest step: $+0.129$ overall, moving exactly the categories routing left behind ($+0.182$ TTL, $+0.181$ VR, $+0.154$ VS). The generated anchor then adds $+0.007$ overall, concentrated where it is triggered ($+0.038$ VR, $+0.032$ VS) and offset by a small cost on provided-image questions ($-0.018$ TTL).
\paragraph{Route, don't fuse.}
Vanilla and + routing differ only in whether lanes are assigned per case or all fired uniformly, and routing wins $0.689$ to $0.680$ while firing strictly fewer lanes.
The per-case lane toggles in Appendix~\ref{sec:appendix_ablations} (Table~\ref{tab:abl_routing}) show why no uniform fusion can win: a lane helps exactly when it carries discriminative signal for its case, and the same lane that adds $+0.04$ in one case subtracts $0.24$ in another. The same logic applies within a case. A hypo-caption describes a single image, so it helps a query that seeks one and cannot help a query that counts them. This is why V-Mem routes the anchor as well as the lanes.

Appendix~\ref{sec:appendix_ablations} reports further ablations, and Appendix~\ref{sec:appendix_errors} shows that on the remaining provided-image errors the evidence is usually retrieved and the answer is still wrong.

\subsection{Efficiency: Build and Answer Cost}
\label{sec:efficiency}

\begin{table}[t]
\centering
\caption{The cost of memory construction and query answering. Lower is better in every row, and the best is in bold.}
\label{tab:efficiency}
\small
\setlength{\tabcolsep}{3.5pt}
\begin{tabular}{@{}llrrrr@{}}
\toprule
Phase & Metric & Omni & M2A & A-Mem & V-Mem \\
\midrule
\multicolumn{6}{@{}l}{\textit{Mem-Gallery}}\\
Build  & LLM tokens   & 2.77M  & 4.3B  & 6.92M         & \textbf{0} \\
Build  & wall-clock   & 6.5\,h & 73\,h & 5.2\,h        & \textbf{120\,s} \\
Answer & tokens / QA  & 2{,}437 & 1.10M & \textbf{959} & 10{,}086 \\
Answer & seconds / QA & 3.31    & 64    & \textbf{0.75} & 2.33 \\
\midrule
\multicolumn{6}{@{}l}{\textit{LoCoMo}}\\
Build  & LLM tokens   & 0.89M   & 133M  & 7.8M          & \textbf{0} \\
Build  & wall-clock   & 1.9\,h  & 11\,h & 5.3\,h        & \textbf{18\,s} \\
Answer & tokens / QA  & \textbf{1{,}783} & 7{,}827 & 2{,}710     & 2{,}302 \\
Answer & seconds / QA & 3.27    & 5.2   & 1.41          & \textbf{1.0} \\
\bottomrule
\end{tabular}
\end{table}

\paragraph{Build is the clean, decisive win.}
Table~\ref{tab:efficiency} reports build and answer cost. Because V-Mem keeps the conversation bound rather than extracting it, its build calls no LLM: 0 tokens, in seconds. The baselines that build their memory with an LLM pay for it before any question is asked: Omni spends around 2.8M tokens and 6.5\,h, A-Mem around 6.9M tokens and 5.2\,h, and M2A's iterative pipeline runs to 4.3B tokens and 73\,h. The cost also recurs as the conversation grows, because new content must be extracted too, while V-Mem's build stays free. The zero-token build is a direct consequence of the retrieval design, not a separate optimization.

\paragraph{Answer cost: a deliberate, modality-aware tradeoff.}
We are explicit that V-Mem does not win every token comparison. On Mem-Gallery it spends more answer-time tokens than the others on visual questions, because every image-bearing query triggers a vision call to extract relevant keywords from the query image, and that is precisely the mechanism that produces its visual-QA wins. The contrast with A-Mem is instructive: A-Mem's answer is the cheapest (959 tokens/QA) exactly because it is text-only and never looks at an image, which is also why it collapses on the visual categories. The vision call can be switched off if the tokens matter more: answers get cheaper and part of the provided-image gain is given up.

\section{Conclusion}
\label{sec:conclusion}

We argued that the difficulty of multimodal long-term memory is not a missing encoder but an unexamined assumption: retrieval by similarity search presumes the most similar content in the memory is the one that answers it, which fails on two independent axes, a \emph{modality gap} and a \emph{similarity-relevance gap}. V-Mem answers both by routing over the (query, target) modality pair and combining two strategies. To cross the modality gap, it matches within the query's own modality and returns the evidence that shares the same round, so it never compares across modalities; and to close the similarity-relevance gap, it searches with an LLM-generated anchor closer to the evidence: a hypothetical caption when a text-only query seeks an image, and an enriched search anchor, the query text plus relevant keywords extracted from the query image, when the query carries an image but seeks text. It returns evidence wrapped in adjacent-round context rather than traversing a graph, and, because the conversation is kept rather than extracted, it builds in seconds with zero LLM tokens. The result is a memory that wins the LLM judge on both a multimodal and a text benchmark, most decisively on the visual questions where caption-only and cross-modal baselines collapse.
\paragraph{Limitations and scope.}
Our evaluation is scoped to image-and-text conversational memory: audio memory is treated as text, but neither benchmark contains audio, and treating audio or video as first-class retrievable modalities with their own lanes remains untested. V-Mem also depends on the backbone's capacity. It reaches evidence by returning the memory content surrounding a match (the matched round together with its neighbors), and leaves the backbone to pick the answer out of that context. A backbone that cannot use that content gains nothing from it, and may even be distracted by it. With a locally served Qwen2.5-VL-7B backbone on Mem-Gallery, V-Mem only scores $0.509$ against Omni's $0.534$, reversing the ordering it holds with a stronger backbone (Appendix~\ref{sec:appendix_local7b}). Finally, the generated anchors are not free. Both add tokens and latency at answer time, and the enriched anchor further needs a vision model, which runs whenever a query carries an image and seeks text. We see this principle, matching through shared rounds and searching with generated anchors, generalizing as multimodal agents accumulate richer, longer memories.

\bibliography{aaai2027}

\appendix

\section{Related Work}
\label{sec:appendix_related}

\paragraph{Memory for LLM Agents.}
Memory is a fundamental component of LLM-based agents~\citep{zhang2025d}, and existing approaches fall into two families. Parametric memory keeps knowledge in the weights, whether by fine-tuning or a built-in memory module~\citep{wang2024b,zhang2025b}. Non-parametric memory instead keeps the interaction history outside the model and retrieves from it, which is where V-Mem and the systems below sit.

MemGPT~\citep{packer2023memgpt} treats the LLM context window as working memory and introduces OS-inspired paging to evict and reload memories from external storage on demand. MemOS~\citep{memos} carries that framing further, treating memory as a resource the system manages in its own right and putting three kinds of memory, parametric, activation, and plaintext, behind a single abstraction that can track them, combine them, and move them between stores. Its concern is how memory is administered over time, not how retrieval behaves when a query and its evidence sit in different modalities. A-Mem~\citep{amem} builds a dynamic note-taking system in which each memory unit is annotated with keywords and linked to related notes, supporting graph-structured text retrieval. SimpleMem~\citep{simplemem} compresses each interaction into a small memory unit that is indexed several ways at once, then at query time works out what the user is looking for before searching, so it can narrow the search rather than scan everything. MAGMA~\citep{magma} splits memory into four separate graphs, semantic, temporal, causal, and entity, and retrieves by learning a policy that walks them, so the route to each retrieved item can be inspected afterwards. Hage~\citep{jiang2026hage} evolves a weighted memory graph through reinforcement learning.

Most of these systems share one commitment: they rewrite the conversation before storing it, by summarizing it, taking notes on it, or building a graph from it. V-Mem departs from that in two ways. First, it is extraction-free: it appends each new exchange and encodes it locally, so building the memory costs zero LLM tokens where these systems spend millions (\S\ref{sec:efficiency}). Second, the multi-hop reasoning that motivates note links and graph edges is served instead by returning each match together with its time-ordered neighbors, which ties with graph-based memory on LoCoMo multi-hop (\S\ref{sec:main_results}).

A system built for multimodal conversations might be expected to give up ground on text-only ones. V-Mem does not: on text-centric LoCoMo it outperforms the text-only agentic baseline on the benchmark that baseline was built for ($0.69$ against $0.51$ for A-Mem). What these systems do share, and what the next subsection takes up, is that they retrieve over text alone, storing textual summaries or sentence embeddings and matching text against text, which leaves multimodal conversations without dedicated support.

\paragraph{Multimodal Agent Memory.}
Pretrained multimodal encoders such as CLIP~\citep{clip}, SigLIP~\citep{siglip}, and BLIP~\citep{li2022blip} project text and images into a shared embedding space for cross-modal matching. While these encoders make modalities comparable, they are trained to align an image with text describing its \emph{visible content}, whereas a memory query about an image typically expresses its \emph{context}: when it was shared, what was discussed, and why it mattered. To address this, MuRAG~\citep{chen2022murag} fine-tunes a multimodal retriever paired with a generation model to retrieve relevant images and text and produce grounded answers; however, it requires task-specific supervision and does not transfer to agentic memory, which has no fixed corpus to train on and keeps growing.

Among agent memory systems that handle multimodal conversations, MemVerse~\citep{memverse} converts all non-text modalities into text chunks, constructs a knowledge graph, and retrieves via graph traversal, leaving raw visual signals out of the search. M2A~\citep{m2a} keeps each image as both an automatically generated caption and an image embedding, and applies iterative retrieval that fuses dense text, sparse lexical, and image signals uniformly, without routing by the (query modality $\times$ target modality) pair. Omni-SimpleMem~\citep{omni} organizes memories into pyramid-structured units and supports multi-granularity retrieval, yet relies on text-based similarity without modeling cross-modal relevance. All three presuppose that the most \emph{similar} retrieved item is the most \emph{relevant} one, an assumption that breaks in two ways in the multimodal long-term memory setting, the modality gap and the similarity-relevance gap (Section~\ref{sec:intro}). Rather than discarding visual information to make retrieval tractable, V-Mem keeps it and gives up the assumption instead: it keeps every image as its own retrievable unit (where MemVerse searches text alone), routes each (query, target) modality pair to the lanes that serve it (where M2A fuses all lanes uniformly), and crosses modalities through round-level binding and generated anchors rather than text similarity alone (where Omni-SimpleMem stops at captions). The effect concentrates exactly where the presupposition fails: on Mem-Gallery's provided-image questions V-Mem reaches $0.872$ where no baseline exceeds $0.472$, and it leads every category under the \texttt{gpt-4o-mini} backbone (\S\ref{sec:main_results}).

\paragraph{Generation-Augmented Retrieval.}
Retrieval can also be improved by generating an intermediate surrogate that narrows the gap between a query and its target evidence, so the two are matched through more similar representations. HyDE~\citep{gao2023hyde} applies this on the query side for retrieving dense text documents: given that a short query and a long passage differ in both volume and format, an LLM expands the query into a hypothetical answer document whose length and style match the target, and the real documents nearest this surrogate are then retrieved without relevance labels. PreMIR~\citep{premir} instead generates on the document side, pre-computing hypothetical queries for every indexed document so that a user query can be matched against them; this moves generation to build time, where its cost scales with the corpus. V-Mem generalizes this query-side, retrieval-time generation to cross-modal long-term memory. When a text-only query seeks an image, an LLM writes a hypothetical caption that \emph{replaces} the query, and the search compares that caption against the stored images' captions. When a query carrying an image seeks text, a vision model extracts relevant keywords from the image and appends them to the query text; the anchor is the resulting \emph{enriched search anchor}, the original query together with those keywords. Either way the anchor is compared against material of its own modality, which is how the modality gap is avoided, and at no indexing cost. Replacing is used selectively, since a hypo-caption describes a single image and cannot serve a query that counts them (Table~\ref{tab:abl_routing}).

\section{Implementation Details}
\label{sec:appendix_impl}

Table~\ref{tab:hyperparams} lists the full V-Mem configuration used in all experiments. All encoders run locally and are used off the shelf, with no fine-tuning. Within each backbone setting, one model at temperature $0$ serves every LLM role at answer time, generating both the anchors and the answer: \texttt{gpt-4o-mini} and the locally served Qwen2.5-VL-32B in the main results, and Qwen2.5-VL-7B in Appendix~\ref{sec:appendix_local7b}. Memory construction calls no LLM at all.

\begin{table}[t]
\centering
\caption{V-Mem hyperparameter configuration. The hypo-caption is skipped on image-counting queries (\S\ref{sec:retrieval}); all routing and gating decisions use the query alone, with no LLM call.}
\label{tab:hyperparams}
\small
\setlength{\tabcolsep}{3pt}
\begin{tabular}{@{}ll@{}}
\toprule
Parameter & Value \\
\midrule
text encoder $\phi$ & \texttt{all-MiniLM-L6-v2} (384-d) \\
image encoder $\psi$ & SigLIP2-base-patch16-384 (768-d) \\
sparse signal & BM25 \\
LLM (anchors $+$ answer) & the answer backbone, temperature 0 \\
\midrule
rank-fusion constant $\kappa$ & 60 \\
per-signal candidate pool & 20 \\
returned rounds $k$ & 12 (30 for LoCoMo aggregation) \\
neighbor window & $\pm1$ \\
image-round filter window & $\pm1$ \\
\midrule
hypo-caption gate & skipped on counting queries \\
image keywords per query & at most 8 \\
\bottomrule
\end{tabular}
\end{table}

\paragraph{Recognizing the query and target modalities.}
V-Mem infers the target modality from the query's wording with a small keyword rule. The target is taken to be an image when the question asks \emph{which} image, photo or picture, or \emph{how many} of them, and to be text otherwise. The rule is deliberately narrow: a question such as ``what is the name in the picture'' mentions an image but wants a fact about it, and is routed to a text target. The query modality is set separately by whether an image is attached. Both decisions use the query alone and call no LLM; the exact patterns are released with the code.

\paragraph{Image-round filter.}
When the target is an image, V-Mem discards any ranked round that holds no image, either in the round itself or in a neighbor within $\pm1$ rounds, and takes the top $k$ from what remains. Such a round has nothing to return: the answer is an image, so a round without one nearby occupies a slot it cannot fill, and text-only rounds that merely mention the subject would otherwise crowd out the rounds that actually hold the picture. The filter applies only to this case, and if it would remove every candidate the unfiltered ranking is kept.

\paragraph{Aggregation questions (LoCoMo only).}
On LoCoMo, V-Mem recognizes aggregation questions from the query's wording with a keyword rule: counting cues such as ``how many'', explicit enumeration cues such as ``list'' or ``in what ways'', and questions asking for a set of items, for example ``what activities has \dots'' or ``what books has \dots''. For these questions V-Mem returns no neighbor rounds and widens the returned set from $12$ rounds to $30$ instead. The branch is not used on Mem-Gallery. As with routing, the decision uses the query alone and calls no LLM; the exact trigger terms are released with the code.

\section{Prompt Library}
\label{sec:appendix_prompts}

V-Mem uses four prompts, all served by the same \texttt{gpt-4o-mini}: an answer prompt, a conflict variant selected when the question asks about a contradiction, and the two anchor-generation prompts. All four are reproduced below, along with the short note appended to the context when the query carries an image, and the judge prompt used for Mem-Gallery. Line breaks and dashes are adjusted to fit the column; the wording is unchanged. Judging uses each benchmark's established prompt unchanged, so our scores stay comparable to published numbers: Mem-Gallery's official LLM-judge prompt, and for LoCoMo the semantic-grader prompt from prior work, whose full text appears in that paper's appendix \citep{magma}.

\paragraph{Answer prompt.}
The default answer prompt. \texttt{\{context\}} is filled with the retrieved rounds and their neighbors, joined into time-ordered blocks as described in \S\ref{sec:retrieval}. When the query carries an image, a short note describing that image is appended to the context; it is shown in the next paragraph.

\begin{lstlisting}[numbers=none, basicstyle=\footnotesize\ttfamily]
You are answering a question using a
person's conversation history.

CONTEXT (each block is one exchange;
some include an image):
{context}

QUESTION: {question}

INSTRUCTIONS:
- Answer ONLY from the context above.
- If the question asks which
  image/photo/picture, answer with the
  image id exactly as shown (e.g.
  D1:IMG_002).
- If the question asks to list or
  enumerate, include EVERY matching
  item, comma-separated.
- Be specific and complete; include
  all details the question asks for.
- Use ONLY what is written in the
  context blocks above -- never
  outside or general knowledge.
- For a yes/no question about whether
  a specific thing was mentioned or
  discussed (e.g. "Did X mention any
  specific ...?"), answer "Not
  mentioned" UNLESS that specific
  thing is explicitly named in the
  context. Do not answer "Yes" based
  on related or general discussion,
  and never invent specifics.
- If the answer is genuinely not in
  the context, reply exactly:
  Information not found

ANSWER:
\end{lstlisting}

\paragraph{Attached-image note.}
The answer call is text-only, so the query's image never reaches the backbone even when the backbone can see images. This note carries the image into the context as text: it states what the attachment shows and, when the image lane returned a match, names the stored image it best matches, so the backbone can identify the entity from the conversation rather than reporting that it found nothing. The middle sentence is present only in that matched case.

\begin{lstlisting}[numbers=none, basicstyle=\footnotesize\ttfamily]
[ATTACHED IMAGE] The user's attached
image shows:
{query_caption}. It visually matches a
conversation image: ({image_id} --
"{caption}"). Identify the
product/entity it depicts using the
conversation and answer; do NOT reply
"Information not found" if the
conversation identifies it.
\end{lstlisting}

\paragraph{Conflict prompt.}
Selected by a keyword check when the question asks about a conflict or contradiction (category CD); it forces a one-word verdict because the references are one word:

\begin{lstlisting}[numbers=none, basicstyle=\footnotesize\ttfamily]
You are checking a person's
conversation history for a conflict or
contradiction. Answer with a clear Yes
or No.

CONTEXT (each block is one exchange):
{context}

QUESTION: {question}

INSTRUCTIONS:
- A CONFLICT exists if the claim (or a
  statement in the conversation)
  asserts a fact that the conversation
  establishes DIFFERENTLY -- e.g. a
  wrong name, attribute, or action for
  an entity the conversation
  describes (the conversation says
  the pet is named Lumi; a claim
  that it is named Coco CONFLICTS),
  or a claim the conversation
  directly contradicts.
- If you identify ANY factual
  discrepancy between the claim and
  what the conversation establishes,
  answer "Yes" -- do NOT explain it
  away as "no conflict".
- For "does X contradict himself / is
  there a contradiction" questions:
  look for two statements in the
  conversation that cannot both be
  true; if such a pair exists, answer
  "Yes".
- Answer "No" ONLY if the claim is
  consistent with / supported by the
  conversation, OR the conversation
  does not address the topic at all
  (an unaddressed topic is NOT a
  conflict).
- Reason internally, but OUTPUT
  EXACTLY ONE WORD: "Yes" or "No" --
  nothing else (no reason, no
  explanation, no punctuation beyond
  the word). The reference answers
  are a single word, so anything more
  is scored wrong.

ANSWER:
\end{lstlisting}

\paragraph{Hypo-caption prompt (text $\to$ image).}
Sent to the text LLM with the question alone; the returned caption replaces the query in the image lane:

\begin{lstlisting}[numbers=none, basicstyle=\footnotesize\ttfamily]
A user is searching the images shared
in a conversation for the ONE image
that answers their question. Write the
caption that image most likely has, in
the style of the conversation's own
captions: a dense noun-phrase
description of what is visible, about
two sentences and 45 words, covering
subjects, attributes, colours,
setting, and any visible text or
labels. Keep whatever the question
names concretely, and add the
appearance it implies but does not
state. Do not begin with "A photo of",
and do not assume the target is a
photograph; it may be an illustration,
an infographic, a screenshot, or a
document.

Question: {q}

Caption:
\end{lstlisting}

\paragraph{Enriched-anchor prompt (image+text $\to$ text).}
Sent to the vision LLM together with the attached image; the returned keywords are appended to the query text to form the anchor both text lanes search:

\begin{lstlisting}[numbers=none, basicstyle=\footnotesize\ttfamily]
You can SEE the user's attached image.

QUESTION: {q}

The answer to that question is in a
past conversation, not in the image.
List the search terms most likely to
appear in the conversation turns that
answer it.

Start from what the question asks
about and find it in the image: if it
asks about a breed, name the animal
and its breed; about a category, name
the item and what kind of thing it is;
about a component, name that part.
Then add the other visible things a
person might have mentioned when
sharing this picture: the main
subject, brands, labels, species,
places, objects.

For each, give the forms a
conversation might use:
the specific name, the everyday name,
and the general category, since it may
use any of them. Include the words the
question itself hinges on.

Give at most 8 terms. Choose them by
one test: how likely is this exact
word to appear in the sentence that
answers the question? Keep terms that
would identify the answer; drop terms
that merely describe the picture.
Single words or short phrases, never
sentences. Prefer names and nouns over
adjectives. Omit anything you are
inferring rather than seeing.

Return JSON: {"keywords": ["...",
"..."]}
\end{lstlisting}

\paragraph{Judge prompt (Mem-Gallery, official).}
The prompt asks for one of five score levels, but the benchmark's scorer collapses them, and every stored score takes one of $\{0, 0.5, 1\}$.
Every system's answers on Mem-Gallery are scored by \texttt{gpt-4o-mini} under the benchmark's released judge prompt, reproduced verbatim:

\begin{lstlisting}[numbers=none, basicstyle=\footnotesize\ttfamily]
You are an impartial judge evaluating
the memory capabilities of an AI
assistant with the question-answering
task. Your task is to compare the
Assistant's Answer against the Ground
Truth and assign a score of 0, 0.25,
0.5, 0.75, or 1.

### Scoring Rubric

**Score 0 (Incorrect / Miss):**
- The answer contradicts the Ground
  Truth.
- For Yes/No questions: The answer has
  the wrong polarity (e.g., says "Yes"
  when Ground Truth is "No").
- For Open-ended questions: The answer
  provides factually wrong information
  or hallucinations.
- The assistant fails to provide the
  required information.

**Score 0.25 (Poor / Tangential):**
- The answer touches on the topic but
  misses the **core entity** or key
  value required.
- The answer contains a mix of minor
  correct details and **significant
  hallucinations** or wrong
  associations.
- The answer is excessively vague to
  the point of being useless (e.g.,
  answering "a dog" instead of "a
  golden retriever").

**Score 0.5 (Partial / Vague):**
- The answer is technically correct,
  but lacks confidence or is
  incomplete.
- The answer captures the **main
  entity or concept** correctly but
  misses a part of the required
  supporting details.
- For Yes/No questions: The polarity
  is correct, but the reasoning is
  flawed (if have), or the assistant
  is uncertain (e.g., "I think it
  might be Yes").
- For Open-ended questions: The answer
  is too general or misses key
  adjectives/details present in the
  Ground Truth.

**Score 0.75 (Good / Minor
Imperfection):**
- The answer is largely accurate and
  captures the core information
  confidently.
- It misses only **minor details**
  (e.g., specific adjectives or
  secondary details) that do not alter
  the main truth.
- The answer contains all the correct
  information but includes unnecessary
  "fluff" or slight conversational
  filler that reduces precision.

**Score 1 (Correct / Exact):**
- The answer is accurate, precise, and
  confident.
- For Yes/No questions: The polarity
  matches the Ground Truth perfectly.
- For Open-ended questions: The answer
  contains **all** the core
  information and necessary details
  required by the Ground Truth without
  hallucinations.

### Input Data

Question: {question} Ground Truth:
{ground_truth} Assistant Answer:
{model_output}

### Output Format

Output strictly in the following JSON
format:
{"score": <0, 0.25, 0.5, 0.75, or 1>,
"reasoning": "<short explanation>"}
\end{lstlisting}

\section{Baseline Configurations and Dataset Statistics}
\label{sec:appendix_baselines}

\paragraph{Protocols.}
We standardized the comparison as follows. Each baseline runs its original implementation with its recommended default settings, and each system answers through its \emph{own} pipeline and prompts, so the comparison is between complete systems rather than retrieval modules grafted onto a shared backbone. Within a backbone setting, four things are held common: the questions; the answering LLM (\texttt{gpt-4o-mini} and the locally served Qwen2.5-VL-32B in the main results, Qwen2.5-VL-7B in Appendix~\ref{sec:appendix_local7b}); the encoders, where a system uses them, since M2A draws on the same sentence encoder, BM25, and SigLIP stack as V-Mem, which isolates the contribution of our routing and anchoring layer; and the judge, \texttt{gpt-4o-mini} under each benchmark's established judge prompt. Everything else is each system's own: Omni is evaluated with its released caption-based Mem-Gallery adapter, faithful to its paper's design, A-Mem runs its official note-construction pipeline, and M2A runs its agentic loop with raw images.

\paragraph{Dataset statistics.}
Table~\ref{tab:dataset_stats} details both benchmarks' question distributions.

\begin{table}[t]
\centering
\caption{Question distributions. Following common practice, LoCoMo's adversarial category is excluded because it rewards abstention rather than retrieval quality.}
\label{tab:dataset_stats}
\small
\setlength{\tabcolsep}{4pt}
\begin{tabular}{@{}llr@{}}
\toprule
Benchmark & Category & Count \\
\midrule
Mem-Gallery & test-time learning (TTL) & 337 \\
(20 dialogues) & visual-centric search (VS) & 306 \\
 & visual-centric reasoning (VR) & 174 \\
 & factual retrieval (FR) & 219 \\
 & multi-entity reasoning (MR) & 206 \\
 & answer refusal (AR) & 184 \\
 & temporal reasoning (TR) & 123 \\
 & knowledge resolution (KR) & 81 \\
 & conflict detection (CD) & 81 \\
 & \textbf{total} & \textbf{1{,}711} \\
\midrule
LoCoMo & single-hop & 841 \\
(10 conversations) & multi-hop & 282 \\
 & temporal & 321 \\
 & open-domain & 96 \\
 & \emph{adversarial (excluded)} & \emph{446} \\
 & \textbf{evaluated} & \textbf{1{,}540} \\
\bottomrule
\end{tabular}
\end{table}

\section{Smaller Backbone Model}
\label{sec:appendix_local7b}

The main results use \texttt{gpt-4o-mini} and a Qwen2.5-VL-32B backbone (Table~\ref{tab:mg_percat}). To test whether V-Mem's advantage survives a weaker answer model, we also ran Mem-Gallery with a locally served Qwen2.5-VL-7B-Instruct (AWQ), holding the retrieval configuration and the \texttt{gpt-4o-mini} judge fixed. As in the 32B setting, each baseline's memory is rebuilt with the backbone that will answer from it, because their construction calls the LLM; V-Mem's build calls no LLM, so it reuses the same memory throughout.

\begin{table}[t]
\centering
\caption{Mem-Gallery with a local Qwen2.5-VL-7B-Instruct (AWQ) backbone, same retrieval configuration and same \texttt{gpt-4o-mini} judge. Best per column in bold, second best underlined.}
\label{tab:local7b}
\small
\begin{tabular}{@{}lrrrr@{}}
\toprule
System & LLM-judge & F1 & BLEU & EM \\
\midrule
Omni (SimpleMem) & \textbf{0.534} & \textbf{0.387} & \textbf{0.141} & \textbf{0.223} \\
V-Mem (ours)     & \underline{0.509} & \underline{0.324} & \underline{0.115} & \underline{0.156} \\
A-Mem            & 0.478 & 0.254 & 0.090 & 0.113 \\
\bottomrule
\end{tabular}
\end{table}

With the smaller backbone V-Mem no longer leads: Omni is ahead on the judge and on every lexical metric, and V-Mem falls to second. V-Mem reaches evidence by returning the memory content surrounding the top match, the matched round together with its neighbors, rather than a unit summarized at build time, so the backbone must locate the answer within that context. A capable backbone benefits from the surrounding content; a smaller one benefits less, and the additional context can instead act as noise that draws it away from the evidence. M2A was not run in this setting. Its memory must be rebuilt for each backbone, and that rebuild already took 219.9 hours under the locally served 32B model (Table~\ref{tab:cost32b}); a third build on local hardware was beyond our compute budget.

\section{Cost under the 32B Backbone}
\label{sec:appendix_cost32b}

Table~\ref{tab:cost32b} reports build and answer cost for the Qwen2.5-VL-32B runs of Table~\ref{tab:mg_percat}. V-Mem's build calls no LLM, so its build cost does not depend on the backbone at all; the baselines run the backbone during construction, so theirs does. A-Mem is cheapest at answer time because it is text-only and never sends an image to the backbone, the same trade-off discussed in \S\ref{sec:efficiency}. Token counts here are not comparable with those in \S\ref{sec:efficiency}, which uses \texttt{gpt-4o-mini}: the two backbones tokenize text differently and charge different numbers of tokens for an image.

\begin{table}[t]
\centering
\caption{Build and answer cost on Mem-Gallery under the Qwen2.5-VL-32B backbone. Lower is better in every row, and the best is in bold.}
\label{tab:cost32b}
\small
\setlength{\tabcolsep}{4pt}
\begin{tabular}{@{}llrrrr@{}}
\toprule
Phase & Metric & Omni & M2A & A-Mem & V-Mem \\
\midrule
Build  & LLM tokens    & 2.97M & 248.2M & 6.01M & \textbf{0} \\
Build  & wall-clock    & 19.9\,h & 219.9\,h & 21.8\,h & \textbf{120\,s} \\
Answer & tokens / QA   & 3{,}553 & 7{,}481 & \textbf{1{,}011} & 4{,}004 \\
Answer & seconds / QA  & 22.7 & 26.3 & \textbf{5.0} & 18.1 \\
\bottomrule
\end{tabular}
\end{table}

\section{Lexical Metrics}
\label{sec:appendix_lexical}

Table~\ref{tab:lexical} complements the LLM-judge results of \S\ref{sec:main_results} with lexical-overlap metrics (F1, BLEU, EM). On Mem-Gallery, V-Mem wins every lexical metric as well. On LoCoMo, Omni wins the lexical metrics while losing the judge. The reason is length: Omni emits long, gold-style answer strings that inflate $n$-gram overlap with the reference, whereas V-Mem answers tersely, and the judge, which scores semantic correctness, prefers V-Mem. V-Mem places third on these metrics, behind A-Mem as well. We therefore treat the LLM judge as the primary metric and report lexical scores for completeness.

Table~\ref{tab:lexical32b} breaks the lexical metrics down by category for both backbones. Under \texttt{gpt-4o-mini} the split is unusually clean: V-Mem leads test-time learning, visual-centric reasoning, visual-centric search, answer refusal and conflict detection on all three metrics, while Omni leads multi-entity reasoning, factual retrieval and temporal reasoning on all three. Knowledge resolution is the only category that divides, with Omni ahead on F1 and BLEU and V-Mem on EM.

Under the Qwen2.5-VL-32B backbone the picture is less uniform: V-Mem leads F1 in six of the nine categories and overall, while Omni leads knowledge resolution and temporal reasoning and A-Mem leads multi-entity reasoning; BLEU follows the same pattern. The two metric families disagree most sharply on A-Mem's visual-search row, whose lexical scores are zero on all three measures while the judge credits $0.449$. A-Mem is text-only and cannot name a stored image by its identifier, so its answers share no surface form with references that are image ids, and the lexical metrics score them at zero whether or not the right picture was identified. The disagreement is therefore about answer format as much as about correctness, and is a reason to read the two metric families together.

\begin{table*}[t]
\centering
\caption{Per-category lexical-overlap metrics on Mem-Gallery under both backbones. Within each block, the best score in each column is in bold. Two provenance notes for the \texttt{gpt-4o-mini} block: V-Mem's lexical scores come from the run preceding the enriched anchor, whose overall judge score is $0.820$ rather than the $0.825$ of Table~\ref{tab:mg_percat}; and Omni's lexical scores are computed over the $1{,}503$ of $1{,}711$ questions for which its stored reports carry lexical values, while its judge scores cover all $1{,}711$.}
\label{tab:lexical32b}
\small
\setlength{\tabcolsep}{3pt}
\begin{tabular}{@{}l rrrrrrrrr r@{}}
\toprule
System & TTL & VR & VS & AR & CD & KR & MR & FR & TR & Overall \\
\midrule
\multicolumn{11}{@{}l}{\textit{\texttt{gpt-4o-mini} backbone, F1}}\\
Omni (SimpleMem) & 0.099 & 0.236 & 0.691 & 0.008 & 0.068 & \textbf{0.290} & \textbf{0.508} & \textbf{0.568} & \textbf{0.542} & 0.356 \\
M2A              & 0.141 & 0.049 & 0.000 & 0.067 & 0.036 & 0.121 & 0.240 & 0.149 & 0.109 & 0.105 \\
A-Mem            & 0.084 & 0.210 & 0.003 & 0.007 & 0.057 & 0.194 & 0.475 & 0.418 & 0.125 & 0.171 \\
\textbf{V-Mem (ours)} & \textbf{0.331} & \textbf{0.426} & \textbf{0.844} & \textbf{0.828} & \textbf{0.571} & 0.267 & 0.485 & 0.467 & 0.440 & \textbf{0.538} \\
\addlinespace[2pt]
\multicolumn{11}{@{}l}{\textit{\texttt{gpt-4o-mini} backbone, BLEU}}\\
Omni (SimpleMem) & 0.046 & 0.062 & 0.227 & 0.001 & 0.008 & \textbf{0.078} & \textbf{0.241} & \textbf{0.234} & \textbf{0.237} & 0.134 \\
M2A              & 0.022 & 0.006 & 0.000 & 0.005 & 0.005 & 0.012 & 0.039 & 0.027 & 0.019 & 0.016 \\
A-Mem            & 0.017 & 0.048 & 0.001 & 0.001 & 0.008 & 0.041 & 0.220 & 0.149 & 0.054 & 0.060 \\
\textbf{V-Mem (ours)} & \textbf{0.075} & \textbf{0.099} & \textbf{0.376} & \textbf{0.259} & \textbf{0.101} & 0.073 & 0.213 & 0.194 & 0.200 & \textbf{0.193} \\
\addlinespace[2pt]
\multicolumn{11}{@{}l}{\textit{\texttt{gpt-4o-mini} backbone, EM}}\\
Omni (SimpleMem) & 0.022 & 0.161 & 0.589 & 0.000 & 0.016 & 0.058 & \textbf{0.055} & \textbf{0.214} & \textbf{0.366} & 0.188 \\
M2A              & 0.000 & 0.011 & 0.000 & 0.000 & 0.012 & 0.000 & 0.000 & 0.000 & 0.000 & 0.001 \\
A-Mem            & 0.000 & 0.149 & 0.003 & 0.000 & 0.025 & 0.000 & 0.015 & 0.027 & 0.008 & 0.023 \\
\textbf{V-Mem (ours)} & \textbf{0.039} & \textbf{0.310} & \textbf{0.670} & \textbf{0.761} & \textbf{0.568} & \textbf{0.062} & 0.029 & 0.073 & 0.301 & \textbf{0.305} \\
\midrule
\multicolumn{11}{@{}l}{\textit{Qwen2.5-VL-32B backbone, F1}}\\
Omni (SimpleMem) & 0.128 & 0.331 & 0.647 & 0.111 & 0.049 & \textbf{0.282} & 0.437 & 0.444 & \textbf{0.515} & 0.349 \\
M2A              & 0.121 & 0.034 & 0.000 & 0.067 & 0.030 & 0.055 & 0.145 & 0.077 & 0.036 & 0.068 \\
A-Mem            & 0.105 & 0.277 & 0.000 & 0.138 & 0.022 & 0.210 & \textbf{0.449} & 0.442 & 0.122 & 0.194 \\
\textbf{V-Mem (ours)} & \textbf{0.258} & \textbf{0.540} & \textbf{0.837} & \textbf{0.273} & \textbf{0.398} & 0.277 & 0.363 & \textbf{0.477} & 0.465 & \textbf{0.455} \\
\addlinespace[2pt]
\multicolumn{11}{@{}l}{\textit{Qwen2.5-VL-32B backbone, BLEU}}\\
Omni (SimpleMem) & 0.054 & 0.076 & 0.228 & 0.007 & 0.008 & \textbf{0.079} & 0.173 & 0.147 & \textbf{0.225} & 0.120 \\
M2A              & 0.015 & 0.005 & 0.000 & 0.004 & 0.004 & 0.004 & 0.014 & 0.009 & 0.007 & 0.008 \\
A-Mem            & 0.029 & 0.059 & 0.000 & 0.008 & 0.003 & 0.045 & \textbf{0.188} & 0.163 & 0.037 & 0.061 \\
\textbf{V-Mem (ours)} & \textbf{0.078} & \textbf{0.116} & \textbf{0.390} & \textbf{0.068} & \textbf{0.070} & 0.078 & 0.119 & \textbf{0.195} & 0.185 & \textbf{0.164} \\
\addlinespace[2pt]
\multicolumn{11}{@{}l}{\textit{Qwen2.5-VL-32B backbone, EM}}\\
Omni (SimpleMem) & 0.065 & 0.259 & 0.529 & 0.000 & 0.037 & 0.086 & \textbf{0.034} & 0.128 & \textbf{0.398} & 0.189 \\
M2A              & 0.000 & 0.000 & 0.000 & 0.000 & 0.012 & 0.000 & 0.000 & 0.000 & 0.000 & 0.001 \\
A-Mem            & 0.015 & 0.218 & 0.000 & 0.000 & 0.012 & 0.025 & 0.029 & 0.155 & 0.016 & 0.051 \\
\textbf{V-Mem (ours)} & \textbf{0.134} & \textbf{0.460} & \textbf{0.722} & \textbf{0.033} & \textbf{0.395} & \textbf{0.099} & 0.024 & \textbf{0.178} & 0.374 & \textbf{0.282} \\
\bottomrule
\end{tabular}
\end{table*}

\begin{table}[t]
\centering
\caption{Lexical-overlap metrics on both benchmarks (LoCoMo excludes the adversarial category, $n{=}1{,}540$). V-Mem wins every lexical metric on Mem-Gallery. On LoCoMo they favor Omni instead, because word overlap rewards length and phrasing regardless of correctness. Best per column in bold, second best underlined.}
\label{tab:lexical}
\small
\setlength{\tabcolsep}{2.5pt}
\begin{tabular}{@{}l rrr c rrr@{}}
\toprule
& \multicolumn{3}{c}{\textbf{Mem-Gallery}} & & \multicolumn{3}{c}{\textbf{LoCoMo}} \\
\cmidrule(lr){2-4}\cmidrule(lr){6-8}
System & F1 & BLEU & EM & & F1 & BLEU & EM \\
\midrule
Omni (SimpleMem) & \underline{0.356} & \underline{0.134} & \underline{0.188} & & \textbf{0.427} & \textbf{0.147} & \textbf{0.183} \\
M2A              & 0.105 & 0.016 & 0.001 & & 0.141 & 0.022 & 0.000 \\
A-Mem            & 0.171 & 0.060 & 0.023 & & \underline{0.311} & \underline{0.097} & \underline{0.088} \\
\textbf{V-Mem (ours)} & \textbf{0.538} & \textbf{0.193} & \textbf{0.305} & & 0.210 & 0.051 & 0.006 \\
\bottomrule
\end{tabular}
\end{table}

\section{Additional Ablations}
\label{sec:appendix_ablations}

\paragraph{Per-case lane toggles.}
Table~\ref{tab:abl_routing} toggles one lane in each routing case of \S\ref{sec:retrieval}. A lane helps exactly when it carries discriminative signal for that case. For text$\to$image, the query text is discriminative, so restricting ranking to image-bearing rounds lifts visual search by $+0.042$. For image+text$\to$image, the query text is generic, so adding the text lane \emph{hurts by $0.24$}: it injects topically related but visually wrong images that displace the true image match under fusion; the image lane alone is best. For image+text$\to$text, the query text alone cannot reach evidence whose wording comes from the attached image, so the anchor folds in the image's keywords. Across all 468 of its questions this is close to a wash, $+0.004$, with 40 questions improving and 35 worsening. We adopt it on design grounds, because the query text alone cannot express what the image contributes, rather than on the strength of this margin. The coexistence of $+$ and $-$ rows across cases means no single uniform fusion can be right everywhere, which is precisely why V-Mem routes lanes and anchors by case. The filter window peaks at $\pm1$ (0.826 / \textbf{0.849} / 0.828 for $\pm0/\pm1/\pm2$), the local image-context radius.

\begin{table*}[t]
\centering
\caption{Routing is forced, not arbitrary. Each row toggles one lane in one routing case. A lane
helps only when it can contribute something the case needs. For text$\to$image, a round holding no image
cannot return one, so restricting the search to image-bearing rounds removes candidates that could never
answer. For image+text$\to$image, the query's own image already carries the search signal, so the text lane
adds nothing the image does not have and instead pulls in rounds matched on words, whose images are the
wrong ones; enabling it costs $-0.24$, and the image lane alone is best. For image+text$\to$text, the
evidence resembles neither the query text nor the query image alone, so the anchor has to combine them, and
adding the image's keywords gives a small gain. A single uniform fusion cannot satisfy both the $+$ and $-$ rows, hence
routing.}
\label{tab:abl_routing}
\small
\begin{tabular}{@{}llrrr@{}}
\toprule
Case (query$\to$target) & Lane ablated & off & on & $\Delta$ \\
\midrule
text$\to$image (VS)  & image-round filter   & 0.807 & \textbf{0.849} & $+0.042$ \\
text$\to$image (VS)  & hypo-caption         & 0.842 & 0.849 & $+0.007$ \\
image+text$\to$image (VS) & text lane (backbone) & \textbf{0.580} & 0.341 & $-0.239$ \\
image+text$\to$text        & image keywords in the anchor & 0.825 & \textbf{0.829} & $+0.004$ \\
\bottomrule
\end{tabular}
\end{table*}

\paragraph{A budget-generous vanilla.}
In the incremental ablation of the main paper (its Table~\ref{tab:abl_incremental}), the first two rungs return bare units while the later rungs return whole rounds, and each round holds several units. The early rungs therefore receive less text than the later ones, which raises a fair objection: perhaps the jump at the round rung is a budget effect rather than a crossing effect. Table~\ref{tab:abl_ladder_budget} tests that by repeating the two unit-level rungs with three times as many units, matching the amount of content a round-level return delivers. Both rungs improve and routing still helps, but the conclusion is unchanged: even with the enlarged budget, the routed unit-level configuration trails the round-level rung by $0.062$ overall and $0.086$ on provided-image questions. Returning more of the most relevant units does not substitute for returning the most relevant rounds.

\begin{table}[t]
\centering
\caption{The two unit-level rungs given three times as many units (Mem-Gallery, LLM judge, $n{=}1{,}711$;
compare the first two rows of the main paper's Table~\ref{tab:abl_incremental}). Even with the larger budget,
the routed unit-level configuration trails the round-level rung by $0.062$ overall. Returning more units
does not substitute for returning whole rounds.}
\label{tab:abl_ladder_budget}
\small
\setlength{\tabcolsep}{2pt}
\begin{tabular}{@{}l rrr r@{}}
\toprule
Configuration & TTL & VR & VS & Overall \\
\midrule
Vanilla, threefold unit budget & 0.835 & 0.420 & 0.681 & 0.745 \\
\;\;+ routing                  & 0.804 & 0.425 & 0.714 & 0.756 \\
\bottomrule
\end{tabular}
\end{table}

\paragraph{Neighbor placement.}
Once the top rounds have been chosen, their neighbors can be added to the context in several ways, and the choice turns out to matter more than one might expect. Table~\ref{tab:abl_granularity} compares four. \emph{Time-ordered runs}, which is what V-Mem does (\S\ref{sec:retrieval}), places each top round together with its $\pm1$ neighbors as one contiguous block in conversation order, and merges blocks that overlap. \emph{Append at tail} keeps the top rounds first and adds all the neighbors after them, so a round and its neighbors end up separated. \emph{No neighbors} returns the top rounds alone. \emph{Interleave and truncate} mixes the neighbors in among the top rounds and then cuts the list back to $k$ entries.

Time-ordered runs are best on average (0.810), ahead of appending at the tail (0.791) and of using no neighbors at all (0.749). Interleave-and-truncate is worst on every dataset (0.729), and its failure is the instructive one: truncating to $k$ evicts top rounds to make room for neighbors, so it gives up recall without delivering the surrounding context it was trying to add. Neighbors must never count against the $k$ budget.

Keeping each round contiguous matters most where an image and the text describing it need to stay together. Merging overlapping windows gains $+0.025$ overall on Mem-Gallery, and the movement is concentrated in the multimodal categories: visual-centric reasoning rises from $0.50$ to $0.75$ and provided-image questions from $0.79$ to $0.86$. It is neutral on the text-only LoCoMo.

\begin{table}[t]
\centering
\caption{How neighbor rounds enter the result matters. Returning each top round as a contiguous,
time-ordered run with its $\pm1$ neighbors (ours) beats, on average, appending neighbors at the tail and
beats using no neighbors. The naive ``interleave then truncate to top-$k$'' is worst on every dataset: it evicts top
rounds to make room for neighbors, losing recall without giving the surrounding context it was adding them
for. Neighbors must never count against the top-$k$ budget.}
\label{tab:abl_granularity}
\small
\begin{tabular}{@{}lrrrr@{}}
\toprule
Neighbor placement & AI-Rob. & Acad. & Fash. & avg \\
\midrule
time-ordered runs (ours)       & 0.754 & \textbf{0.831} & \textbf{0.844} & \textbf{0.810} \\
append at tail                 & \textbf{0.772} & 0.773 & 0.828 & 0.791 \\
no neighbors ($\pm0$)          & 0.728 & 0.714 & 0.803 & 0.749 \\
interleave + truncate          & 0.684 & 0.708 & 0.795 & 0.729 \\
\bottomrule
\end{tabular}
\end{table}

\paragraph{Caption folding.}
Mem-Gallery supplies a caption with every shared image, and V-Mem can store it two ways: attached to the image alone, or additionally written into the text of the turn in which the image was shared, so that it is indexed like any other conversation text. The second is what we call \emph{folding}, and it is what V-Mem does (\S\ref{sec:construction}).

Folding improves the LLM judge from 0.780 to 0.800 (weighted, $n{=}195$) and never hurts a category. The gains sit where caption text gives the text lane something concrete to match: visual-centric reasoning $+0.083$, knowledge resolution $+0.071$, temporal reasoning $+0.036$. Image-driven categories are unchanged, since image-to-image matching never consults caption text. Folding costs nothing at build time, because the caption is already supplied by the dataset and only has to be copied into the turn.

\paragraph{The image encoder.}
V-Mem uses its image encoder almost entirely to match one image against another, which raises a natural question: is a contrastive image-text model the right tool for that, or would an encoder trained purely on images do better? The two differ in what they were trained for. SigLIP learns to align an image with text describing it, so its embedding reflects what an image is \emph{about}. DINOv2 is self-supervised on images alone and is built for visual similarity, so its embedding reflects what an image \emph{looks like}.

Table~\ref{tab:abl_encoder} swaps one for the other over the full benchmark, with both arms measured under the same configuration so the comparison is paired. The overall judge barely moves, $0.820$ to $0.817$, and the per-category pattern follows those training objectives: DINOv2 is slightly better at recognizing the same object again (test-time learning, visual-centric reasoning), while SigLIP is better when the query describes what to look for (visual-centric search, $+0.029$). The two effects cancel. A purpose-built image encoder therefore does not improve image-to-image matching at this scale, so we keep SigLIP, which has the further advantage of matching the encoder stack the baselines use.

\begin{table}[t]
\centering
\caption{A contrastive image encoder is adequate for image-to-image matching. Swapping SigLIP (trained to
align images with text) for DINOv2 (self-supervised on images alone) over the full benchmark leaves the
overall judge unchanged ($-0.003$, within noise). DINOv2 is slightly better at recognizing the same object
again (TTL, VR) while SigLIP is better at finding an image from a description (VS, $+0.029$), and the two
cancel. Only image-to-image cases are affected; text categories are unchanged by construction. Both arms
were measured on the configuration preceding the enriched anchor, so the SigLIP column reads $0.820$ overall
rather than the $0.825$ of Table~\ref{tab:mg_percat}.}
\label{tab:abl_encoder}
\small
\begin{tabular}{@{}l rr c@{}}
\toprule
Category & SigLIP & DINOv2 & $\Delta$ \\
\midrule
TTL (instance)        & 0.874 & \textbf{0.875} & $+0.002$ \\
VR  (visual reason.)  & 0.569 & \textbf{0.578} & $+0.009$ \\
VS  (visual search)   & \textbf{0.827} & 0.797 & $-0.029$ \\
\midrule
\textbf{Overall judge} & \textbf{0.820} & 0.817 & $-0.003$ \\
\bottomrule
\end{tabular}
\end{table}

\clearpage
\section{Statistical Significance}
\label{sec:appendix_significance}

Every score in this paper is a mean over per-question LLM-judge outcomes, and all systems answer the same questions, so system differences are paired. We quantify them two ways, both computed from the stored judge outputs without re-running any system. First, a paired bootstrap over questions: the question set is resampled with replacement $10{,}000$ times, each system's mean is recomputed on each resample, and the $2.5$ and $97.5$ percentiles of the resampled difference give a $95\%$ confidence interval. Second, a two-sided sign-flip permutation test on the paired per-question differences ($10{,}000$ permutations), which makes no parametric assumption about the score distribution. On Mem-Gallery V-Mem's overall score is $0.825$ with a $95\%$ interval of $[0.808, 0.841]$; on LoCoMo it is $0.690$ with $[0.672, 0.710]$.

\begin{table}[htb]
\centering
\caption{Paired bootstrap and sign-flip permutation tests on the per-question judge scores ($10{,}000$ resamples each). $\Delta$ is the paired mean difference, the interval is its $95\%$ bootstrap confidence interval, and $p$ is two-sided from the permutation test. In the ablation block each row compares a rung with the rung immediately above it.}
\label{tab:significance}
\small
\setlength{\tabcolsep}{3pt}
\begin{tabular}{@{}lrcr@{}}
\toprule
Comparison & $\Delta$ & 95\% CI & $p$ \\
\midrule
\multicolumn{4}{@{}l}{\textit{Mem-Gallery} ($n=1{,}711$)}\\
V-Mem vs.\ Omni   & $+0.263$ & $[+0.240, +0.286]$ & $<10^{-4}$ \\
V-Mem vs.\ A-Mem  & $+0.365$ & $[+0.342, +0.388]$ & $<10^{-4}$ \\
V-Mem vs.\ M2A    & $+0.511$ & $[+0.488, +0.534]$ & $<10^{-4}$ \\
\midrule
\multicolumn{4}{@{}l}{\textit{LoCoMo} ($n=1{,}540$)}\\
V-Mem vs.\ Omni   & $+0.108$ & $[+0.086, +0.129]$ & $<10^{-4}$ \\
V-Mem vs.\ A-Mem  & $+0.179$ & $[+0.157, +0.200]$ & $<10^{-4}$ \\
V-Mem vs.\ M2A    & $+0.269$ & $[+0.245, +0.293]$ & $<10^{-4}$ \\
\midrule
\multicolumn{4}{@{}l}{\textit{Incremental ablation} (Mem-Gallery, $n=1{,}711$)}\\
$+$ routing            & $+0.009$ & $[-0.003, +0.021]$ & $0.16$ \\
$+$ shared rounds      & $+0.129$ & $[+0.111, +0.148]$ & $<10^{-4}$ \\
$+$ generated anchor   & $+0.007$ & $[-0.002, +0.015]$ & $0.14$ \\
\bottomrule
\end{tabular}
\end{table}

Table~\ref{tab:significance} reports the results. Every margin over every baseline, on both benchmarks, is significant well beyond the conventional threshold, and the smallest of them, $+0.108$ over Omni on LoCoMo, has a confidence interval whose lower bound still sits at $+0.086$.

The ablation ladder is a different matter, and we state it plainly: only one of its three steps is significant on the overall score. Matching through shared rounds gains $+0.129$ at $p<10^{-4}$, while routing ($+0.009$, $p=0.16$) and the generated anchor ($+0.007$, $p=0.14$) both have intervals that include zero. Neither step is defended here on its overall margin. Routing is justified by the per-case lane toggles of Table~\ref{tab:abl_routing}, where a single lane swings a case by $-0.24$, and the generated anchor by the per-category gains it produces where it fires ($+0.038$ VR, $+0.032$ VS). Overall means average those effects away against the categories a step does not touch.

\section{Algorithm and Reproducibility}
\label{sec:appendix_algorithm}

\paragraph{The retrieval procedure.}
Algorithm~\ref{alg:vmem} states routed retrieval in full. Its shape is the same in every case: the case is read from the query alone, that case fixes two things, which units are searchable ($\mathcal{U}$) and what each active lane searches with (the anchors $A$), and everything after those two choices is common.

The four branches differ only in those choices. For image+text$\to$image the searchable set is the stored images and the sole anchor is the query image, so no text lane runs at all. For text$\to$image the search is confined to image-bearing rounds, the text lane keeps the query, and the image lane is anchored on the generated caption $\hat c$. For image+text$\to$text a vision model extracts relevant keywords $\hat w$ from the query image, the text anchor becomes the query text with those keywords appended, and the image lane runs alongside it. For text$\to$text the text lane searches all units with the raw query.

What follows is shared. Each active lane scores every candidate unit against its own anchor, the lanes are fused by reciprocal rank, each round takes the score of its best-matching unit, and the top $k$ rounds are returned together with their $\pm W$ neighbors, merged into time-ordered blocks. Aggregation questions are the one exception: they drop the neighbors and widen $k$ instead. The update operator of Eq.~(\ref{eq:mag}) is the zero-LLM construction of \S\ref{sec:construction}.

\begin{algorithm}[t]
\caption{V-Mem Routed Retrieval. Routing fixes what to search \emph{over} ($\mathcal{U}$), what to search \emph{with} (anchors $A$), and what to \emph{return} ($\Pi$).}
\label{alg:vmem}
\begin{algorithmic}[1]
\REQUIRE query $q=(q^{\mathsf{txt}},q^{\mathsf{img}})$; encoders $\phi,\psi$; RRF constant $\kappa$; round budget $k$; neighbor window $W$
\STATE $c \leftarrow \mathrm{Route}(q)$, the case $(m(q),m(u^{\star}))$ inferred from the query alone
\IF{$c = \mathsf{img}\!+\!\mathsf{txt}\!\rightarrow\!\mathsf{img}$}
    \STATE $\mathcal{U} \leftarrow$ images;\quad $A \leftarrow \{\text{image lane } \psi\!:\, q^{\mathsf{img}}\}$
\ELSIF{$c = \mathsf{txt}\!\rightarrow\!\mathsf{img}$}
    \STATE $\hat c \leftarrow \mathrm{LLM}(q^{\mathsf{txt}})$;\quad $\mathcal{U} \leftarrow$ image-bearing rounds
    \STATE $A \leftarrow \{\text{text lane}\!:\, q^{\mathsf{txt}};\ \text{image lane (caption)}\!:\, \hat c\}$
\ELSIF{$c = \mathsf{img}\!+\!\mathsf{txt}\!\rightarrow\!\mathsf{txt}$}
    \STATE $\hat w \leftarrow \mathrm{LLM}(q^{\mathsf{txt}},q^{\mathsf{img}})$ \quad(relevant keywords extracted from the query image)
    \STATE $\mathcal{U} \leftarrow$ all units;\quad $A \leftarrow \{\text{image lane } \psi\!:\, q^{\mathsf{img}};\ \text{text lane}\!:\, q^{\mathsf{txt}}\!\parallel\!\hat w\}$
\ELSE
    \STATE $\mathcal{U} \leftarrow$ all units;\quad $A \leftarrow \{\text{text lane}\!:\, q^{\mathsf{txt}}\}$ \quad($c = \mathsf{txt}\!\rightarrow\!\mathsf{txt}$)
\ENDIF
\STATE for each lane $\ell$ with anchor $a \in A$: score $u \in \mathcal{U}$ by $s_\ell(a,u)$; fuse lanes by RRF$(\kappa)$
\STATE $\mathrm{score}(r) \leftarrow \max_{u \in r} \mathrm{fused}(u)$;\quad $R \leftarrow$ top-$k$ rounds
\RETURN $\Pi(R)$: each round with its $\pm W$ neighbors, merged and time-ordered (aggregation: drop neighbors, widen $k$)
\end{algorithmic}
\end{algorithm}

\paragraph{Reproducibility checklist.}
We summarize the elements needed to reproduce our results; the configuration is fixed across all
experiments (\S\ref{sec:setup}, Appendix~\ref{sec:appendix_impl}).
\begin{itemize}
\item \textbf{Datasets.} Mem-Gallery~\citep{memgallery} ($1{,}711$ questions, 20 dialogues) and
LoCoMo~\citep{locomo} ($1{,}540$ evaluated questions, 10 conversations); both are public. Full
category counts are in Table~\ref{tab:dataset_stats}.
\item \textbf{Models and encoders.} Text $\phi$: \texttt{all-MiniLM-L6-v2} (384-d); image $\psi$:
SigLIP2-base-patch16-384 (768-d); sparse: BM25; all run locally, off the shelf, with no fine-tuning.
Within each backbone setting one model at temperature $0$ serves both the anchor-generation and
answer roles, \texttt{gpt-4o-mini} or the local Qwen2.5-VL model, while the judge is always
\texttt{gpt-4o-mini}; construction calls no LLM.
\item \textbf{Hyperparameters.} All values are fixed across every experiment and listed in
Table~\ref{tab:hyperparams}, with no per-dataset tuning. The one dataset-dependent element is the
aggregation branch of Appendix~\ref{sec:appendix_impl}, which fires only on LoCoMo, is triggered by
the query's wording, and was not tuned against results. The rank-fusion constant $\kappa{=}60$ is the standard RRF default, and the candidate
pool ($20$) and returned rounds ($k{=}12$) are conventional retrieval budgets, held fixed as
controlled variables rather than tuned. The neighbor window ($\pm1$) was chosen against the
alternatives in Table~\ref{tab:abl_granularity}.
\item \textbf{Evaluation.} Primary metric is an LLM-as-judge score from \texttt{gpt-4o-mini} under each
benchmark's established prompt (Mem-Gallery's official prompt; LoCoMo's from \citep{magma}); lexical
F1/BLEU/EM are secondary (Appendix~\ref{sec:appendix_lexical}). Every question is evaluated once; the
answer LLM and judge run at temperature $0$. Differences between systems are tested with a paired
bootstrap and a sign-flip permutation test over questions (Appendix~\ref{sec:appendix_significance}).
\item \textbf{Baselines.} Each baseline uses its original implementation and recommended settings,
with the shared answer LLM, encoders (where applicable), and judge described in
Appendix~\ref{sec:appendix_baselines}.
\item \textbf{Prompts.} All V-Mem prompts are reproduced verbatim in Appendix~\ref{sec:appendix_prompts}.
\item \textbf{Code and compute.} All code, prompts, and configuration files will be released
publicly. Two machines are used. A workstation with a 24-core Intel i9-13900, 128\,GB RAM, and one
NVIDIA RTX A2000 (12\,GB) runs the encoders $\phi$ and $\psi$, the memory build, and the
Qwen2.5-VL-7B backbone of Appendix~\ref{sec:appendix_local7b}. A server with two NVIDIA RTX A5000
(24\,GB each), a 20-core Intel Xeon Silver 4210R, and 628\,GB RAM under Ubuntu 20.04 and CUDA 12.2
serves Qwen2.5-VL-32B, quantized to Q4\_K\_M, sharded across both GPUs under Ollama with a
$40{,}960$-token context and the model held resident between calls. \texttt{gpt-4o-mini} is reached
through a hosted API. The build costs in Table~\ref{tab:efficiency} are wall-clock on the
workstation; those in Table~\ref{tab:cost32b} are wall-clock for the same pipelines driven against
the server-hosted 32B model. The per-question answer times include API latency.
\end{itemize}

\section{Case Study}
\label{sec:appendix_case}

Table~\ref{tab:case_study} traces one real Mem-Gallery question per mechanism, comparing the full system against the same system with that mechanism disabled (the corresponding rung of the incremental ablation, \S\ref{sec:ablations}). The three cases mirror Figure~\ref{fig:motivation} with benchmark data.

\emph{Case 1, matching through shared rounds.} The query image finds the right stored photo, but the exchange carrying that photo says only that it is being shared; the catalog category is named one exchange earlier. Unit-level retrieval returns the matched image bare and the backbone answers from an unrelated product unit; round-level return with neighbors carries the naming exchange into context.

\emph{Case 2, hypo-caption.} The query's only content word (``friends'') appears in no stored caption, so the raw anchor retrieves nothing and the system abstains. The drafted caption describes the target scene in caption vocabulary, and the correct image wins.

\emph{Case 3, enriched anchor.} The question asks which entrepreneur was discussed alongside the attached picture, and names nobody. The picture shows a Tesla, and names nobody either. The evidence therefore resembles neither part alone, and searching with the query text returns an image id where a person's name was wanted. The vision model reads the terms visible in the picture, \emph{Tesla} among them, and appending those to the question reaches the turn that discusses Elon Musk.

\begin{table*}[t]
\centering
\caption{Case study: one real benchmark question per mechanism. Each row shows what the system answers with the mechanism disabled and with it enabled. Questions and correct answers are quoted from Mem-Gallery.}
\label{tab:case_study}
\small
\setlength{\tabcolsep}{4pt}
\begin{tabular}{@{}p{0.30\textwidth}p{0.29\textwidth}p{0.33\textwidth}@{}}
\toprule
Question and correct answer & Without the mechanism & With the mechanism \\
\midrule
``According to the conversation, what category is the clothing in the picture?'', with a photo attached. \newline \emph{Correct:} Collared Luxe Faux Fur Jacket &
\emph{Returning bare units:} \newline ``Cowl Neck Fisherman Sweaters'' &
\emph{Returning whole rounds:} \newline ``The clothing in the picture is categorized as a `Collared Luxe Faux Fur Jacket.'\,'' \\
\midrule
``Which picture shown in the conversation is about friends?'' \newline \emph{Correct:} D12:IMG\_002 &
\emph{Searching with the query:} \newline ``Information not found'' &
\emph{Searching with a hypo-caption:} \newline ``D12:IMG\_002'' \\
\midrule
``When Alex was talking about which entrepreneur did he share an image related to this picture?'', with a photo of a Tesla. \newline \emph{Correct:} Elon Musk &
\emph{Searching with the query:} \newline ``The image related to the entrepreneur discussed by Alex is D1:IMG\_002.'' &
\emph{Searching with an enriched anchor:} \newline ``Alex shared an image related to Elon Musk, specifically about Tesla\ldots'' \\
\bottomrule
\end{tabular}
\end{table*}

\section{Error Analysis: Provided-Image Questions}
\label{sec:appendix_errors}

The remaining errors on provided-image questions are not retrieval failures: the evidence is there and the answer is still wrong. We audited the provided-image misses of the round-level configuration: in all 21, the returned context contained the conversation text naming the answer, and the answer model did not use it. The pattern is consistent. The answer call is text-only, so what the model sees is the attached-image note describing the picture (Appendix~\ref{sec:appendix_prompts}) alongside a conversation that explicitly identifies what it depicts. It follows the description and overrides the conversation, answering \emph{``Blue Mosque''} where the conversation says Hagia Sophia, or \emph{``Cocker Spaniel''} where it says Brittany Spaniel.

Prompting did not fix it. Instructing the model to treat that description as a guess from pixels recovered almost none of the misses while breaking many answers that had been correct, taking the provided-image judge from $0.895$ to $0.745$ across the 468 questions of this case. The same confidently asserted description that causes these errors is also the main route to the answers the model already gets right, because it frequently names the answer entity outright, so suppressing it costs more than it saves. We therefore report this as a limitation of the answer model rather than patch around it. Reconciling the attached-image description with the retrieved statements is future work.

\end{document}